\pdfoutput=1
\PassOptionsToPackage{table}{xcolor}
\documentclass[11pt]{alta}

\usepackage[round]{natbib}

\usepackage{amsmath,amsfonts,bm}

\def\eqref#1{equation~\ref{#1}}

\def\1{\bm{1}}

\DeclareMathAlphabet{\mathsfit}{\encodingdefault}{\sfdefault}{m}{sl}
\SetMathAlphabet{\mathsfit}{bold}{\encodingdefault}{\sfdefault}{bx}{n}

\usepackage{array}
\usepackage{tabularx}
\usepackage{graphicx}
\usepackage{subcaption}        
\usepackage{booktabs}
\usepackage{multirow}
\usepackage{wrapfig}
\usepackage{enumitem}          
\usepackage{algorithm}         
\usepackage{algpseudocode}
\algrenewcommand{\algorithmiccomment}[1]{\hfill{\color{gainink}\footnotesize\itshape // #1}}

\usepackage{url}
\usepackage{xcolor}
\usepackage{colortbl}
\definecolor{grouptint}{HTML}{F6F7F9}
\definecolor{besttint}{HTML}{dee2ee}  
\definecolor{metatint}{HTML}{E9EFF9}  
\definecolor{coldtint}{HTML}{E6DCF0}  
\definecolor{evotint}{HTML}{EAD0EC}   
\definecolor{gainink}{RGB}{46,125,50}
\definecolor{lossink}{RGB}{166,64,58}
\definecolor{deltagrey}{RGB}{110,110,110}
\newcommand{\gain}[1]{\textcolor{gainink}{#1}}
\newcommand{\twoline}[2]{\begin{tabular}[c]{@{}c@{}}#1\\#2\end{tabular}} 
\newcommand{\loss}[1]{\textcolor{lossink}{#1}}

\tcbuselibrary{most}
\definecolor{takeawayblue}{HTML}{1F4E79}  
\newtcolorbox[auto counter]{takeaway}{%
  enhanced, breakable,
  colback=takeawayblue!4, colframe=takeawayblue!85,
  colbacktitle=takeawayblue!85, coltitle=white,
  boxrule=0.5pt, arc=2pt,
  left=6pt, right=6pt, top=4pt, bottom=4pt,
  fonttitle=\bfseries\small,
  title={Takeaway~\thetcbcounter}}

\definecolor{casegray}{RGB}{100,113,124}
\tcbset{casebase/.style={%
  enhanced, colback=white, boxrule=0.5pt, arc=1.5pt,
  left=3pt, right=3pt, top=2pt, bottom=2pt, middle=2pt, boxsep=1pt,
  segmentation style={dashed, line width=0.4pt},
  fonttitle=\bfseries\footnotesize, fontupper=\footnotesize, fontlower=\footnotesize,
  before upper={\raggedright}, before lower={\raggedright}}}
\newtcolorbox{casebox}[1]{casebase, colframe=casegray, colbacktitle=casegray!12, coltitle=black,
  segmentation style={casegray!60, dashed, line width=0.4pt}, title={#1}}
\newtcolorbox{metacasebox}[1]{casebase, colframe=takeawayblue!85, colbacktitle=takeawayblue!85, coltitle=white,
  segmentation style={takeawayblue!50, dashed, line width=0.4pt}, title={#1}}
\definecolor{Instruction followingfill}{RGB}{253,235,217}  \definecolor{Instruction followingacc}{RGB}{204,112,52}
\definecolor{Context managementfill}{RGB}{251,228,234} \definecolor{Context managementacc}{RGB}{190,84,116}
\definecolor{Tool executionfill}{RGB}{221,243,238} \definecolor{Tool executionacc}{RGB}{36,140,122}
\definecolor{Verificationfill}{RGB}{230,225,246} \definecolor{Verificationacc}{RGB}{112,88,184}
\usepackage{pifont}
\newcommand{\micon}[1]{\raisebox{-0.45ex}{\includegraphics[height=2.2ex]{figure/icons/#1.png}}\,}
\newcommand{\okmark}{\textcolor{gainink}{\ding{51}}}
\newcommand{\badmark}{\textcolor{lossink}{\ding{55}}}
\tcbset{casestep/.style={%
  enhanced, boxrule=0pt, leftrule=1.5pt, arc=1pt,
  left=3pt, right=14pt, top=1.5pt, bottom=1.5pt, boxsep=0.5pt,
  before skip=3pt, after skip=3pt, fontupper=\footnotesize,
  before upper={\raggedright}}}
\expandafter\def\csname caseLabelTool execution\endcsname{Execution}
\expandafter\def\csname caseLabelContext management\endcsname{Context}
\expandafter\def\csname caseLabelInstruction following\endcsname{Instruction}
\expandafter\def\csname caseLabelVerification\endcsname{Verification}
\newtcolorbox{metabox}[2]{casestep, colback={#1fill}, colframe={#1acc},
  overlay={\node[anchor=north east, inner sep=0pt] at ([xshift=-3pt, yshift=-3pt]frame.north east) {\footnotesize #2};},
  before upper={\raggedright{\tiny\bfseries\textcolor{#1acc}{[\csname caseLabel#1\endcsname]}}\hspace{2pt}}}
\newtcolorbox{plainbox}[1]{casestep, colback=casegray!8, colframe=casegray!60,
  overlay={\node[anchor=north east, inner sep=0pt] at ([xshift=-3pt, yshift=-3pt]frame.north east) {\footnotesize #1};}}
\newcommand{\caseh}[1]{{\bfseries #1}}
\newcommand{\shown}[1]{{\setlength{\fboxsep}{1pt}\setlength{\fboxrule}{0.5pt}\fcolorbox{casegray}{white}{#1}}}
\newcommand{\casereward}[3]{{\footnotesize\textbf{Reward:}~r0~#1~~$\vert$~~r1--r5~#2\quad #3}}
\newcommand{\casetitle}[2]{#1\hfill{\normalfont\footnotesize #2}}

\definecolor{AltaPrimary}{HTML}{1F4E79}  
\definecolor{AltaLink}{HTML}{2A6F97}    
\definecolor{AltaAmber}{HTML}{C96A86}   
\definecolor{AltaTint}{HTML}{F1F4F8}    

\title{Evolving to Match Capabilities: Improving Harness Self-Evolution via Skill Adaptation}
\author{Zhihao Yang\textsuperscript{1,7,*}\and
  Haojin Wang\textsuperscript{2,*}\and
  Juncheng Wu\textsuperscript{3,*}\and
  Kuan Zhang\textsuperscript{4}\and
  Yukun Chen\textsuperscript{1}\and
  Run Luo\textsuperscript{5}\\
  Yaojie Lu\textsuperscript{6}\and
  Yuyin Zhou\textsuperscript{3}\and
  Lu Wang\textsuperscript{4,\dag}\and
  Xianpei Han\textsuperscript{6,\dag}}
\affiliation{\textsuperscript{1}University of Chinese Academy of Sciences\quad
  \textsuperscript{2}Independent Researcher\quad
  \textsuperscript{3}University of California, Santa Cruz\\
  \textsuperscript{4}Tsinghua University\quad
  \textsuperscript{5}National University of Singapore\quad
  \textsuperscript{6}Chinese Information Processing Laboratory, Institute of Software, Chinese Academy of Sciences\quad
  \textsuperscript{7}School of Advanced Interdisciplinary Sciences, University of Chinese Academy of Sciences\\
  \textsuperscript{*}Equal contribution\quad
  \textsuperscript{\dag}Corresponding authors\\
  \texttt{yangzhihao25@mails.ucas.ac.cn}\\[6pt]
  {\small
    \textbf{Project Page:}\enspace
    \href{https://zhihaoyang30.github.io/skill-adaptation/}
         {\textcolor{AltaLink}
           {\texttt{zhihaoyang30.github.io/skill-adaptation}}}
  }
}
\runningtitle{Evolving to Match Capabilities}
\newcommand{\wordmark}[1]{\raisebox{-0.5\height}{\includegraphics[height=17pt]{figures/#1-logo.png}}}
\newcommand{\crest}[1]{\raisebox{-0.5\height}{\includegraphics[height=25pt]{figures/#1-logo.png}}}
\newcommand{\logogap}{\hspace{15pt}}
\newcommand{\logostrip}{%
  \wordmark{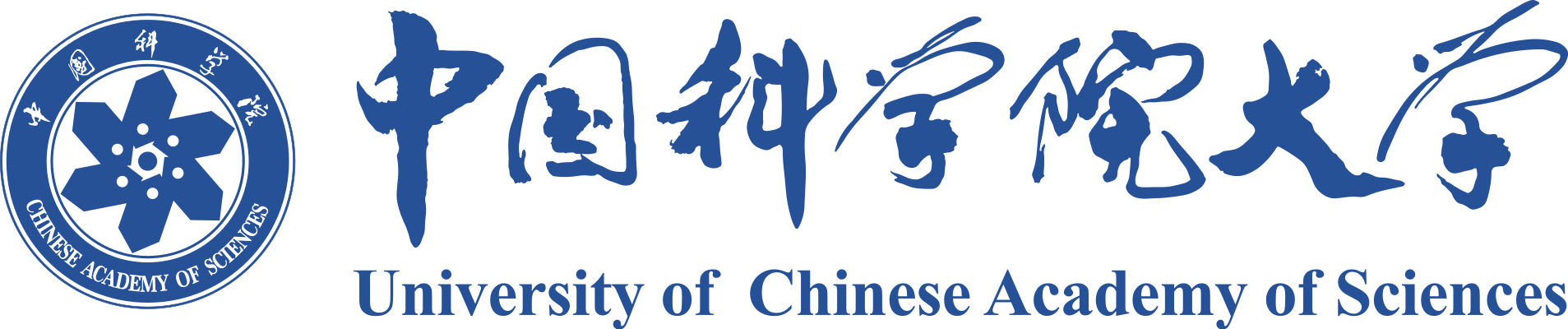}\logogap
  \wordmark{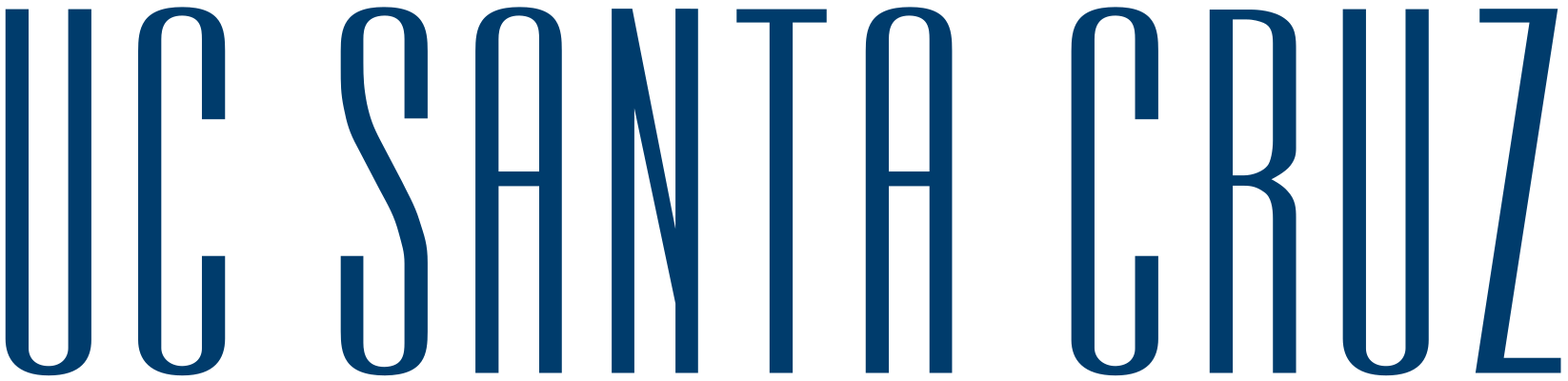}\logogap
  \crest{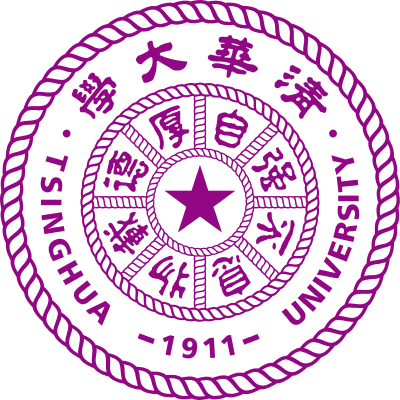}\logogap
  \crest{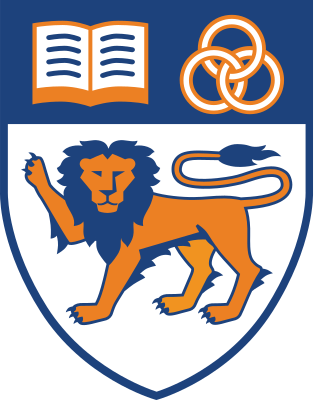}}
\fancypagestyle{altafirst}{%
  \fancyhf{}%
  \fancyhead[C]{\logostrip}%
  \fancyfoot[R]{\sffamily\fontsize{8}{10}\selectfont\color{AltaGrey}\thepage}%
  \renewcommand{\headrulewidth}{0pt}}

\hypersetup{pdftitle={Evolving to Match Capabilities: Improving Harness Self-Evolution via Skill Adaptation},
            pdfauthor={Zhihao Yang, Haojin Wang, Juncheng Wu, Kuan Zhang, Yukun Chen, Run Luo, Yaojie Lu, Yuyin Zhou, Lu Wang, Xianpei Han}}

\usepackage[capitalise,nameinlink]{cleveref}  

\crefname{section}{\S\!}{\S\S\!}
\crefname{table}{Tab.}{Tabs.}
\crefname{figure}{Fig.}{Figs.}
\crefname{algorithm}{Alg.}{Algs.}
\crefname{appendix}{App.}{Apps.}
\crefname{equation}{Eq.}{Eqs.}

\begin{document}
\begin{abstract}

Harness self-evolution offers a practical route for LLM agents to improve from execution experience without updating model parameters.
Its effectiveness depends on whether models can successfully act on the guidance provided by their harnesses. 
However, we find that harness benefits are not uniform across model capabilities: skills that improve stronger models can provide little benefit or even degrade performance for weaker ones.
Across seven models and three benchmarks, we identify 103 task–skill pairs where skills improve GPT-5.6-luna by 31.1 points but degrade Gemma-4-E4B by 8.8 points.
In 86\% of failed weak-model runs, the violated requirement is already stated in the skill. This motivates skill adaptation: reformulating existing skills to match the executor’s capabilities and make their guidance easier to execute. 
To achieve this, we diagnose failures in tool execution, context management, instruction following, and verification, and distill these diagnoses into a fixed meta-skill that guides adaptation during harness self-evolution.
Guided by the meta-skill, a capable external rewriter, Kimi-K3, adapts skills without execution evidence, improving the three weak executors by 5.1–14.1 points over the original skills. We then incorporate this adaptation into a five-round self-evolution loop, where each model revises its own skill from execution trajectories and verifier feedback. Across seven executors, from Gemma-4-E4B to Kimi-K2.5, the meta-skill consistently improves final scores by 2.0–5.1 points over the same loop without the meta-skill.
Our findings provide practical design guidance for developing more effective harness self-evolution methods.
\end{abstract}

\maketitle

\section{Introduction}
\label{sec:intro}

Recent advancements in modern Large Language Model (LLM) agents have shifted from a static prompt-response generation to dynamic agentic workflows~\citep{yao2022react,wang2023voyager, team2026kimi}.
With the emergence of agentic workflows, LLM agents increasingly rely on external harnesses to aid in the execution of tasks~\citep{yang2024swe, merrill2026terminal,li2026skillsbench}.
Harness self-evolution automates the construction and refinement of these harnesses using execution trajectories and feedback while keeping model parameters fixed~\citep{weng2026harness,lin2026harness,lou2026autoharness}. This enables agents to develop harnesses suited to different downstream tasks, reducing the need for manual harness design and maintenance.

\begin{figure*}[!t]
\centering
\begin{subfigure}{0pt}\phantomsubcaption\label{fig:teaser-a}\end{subfigure}%
\begin{subfigure}{0pt}\phantomsubcaption\label{fig:teaser-b}\end{subfigure}%
\includegraphics[width=\textwidth]{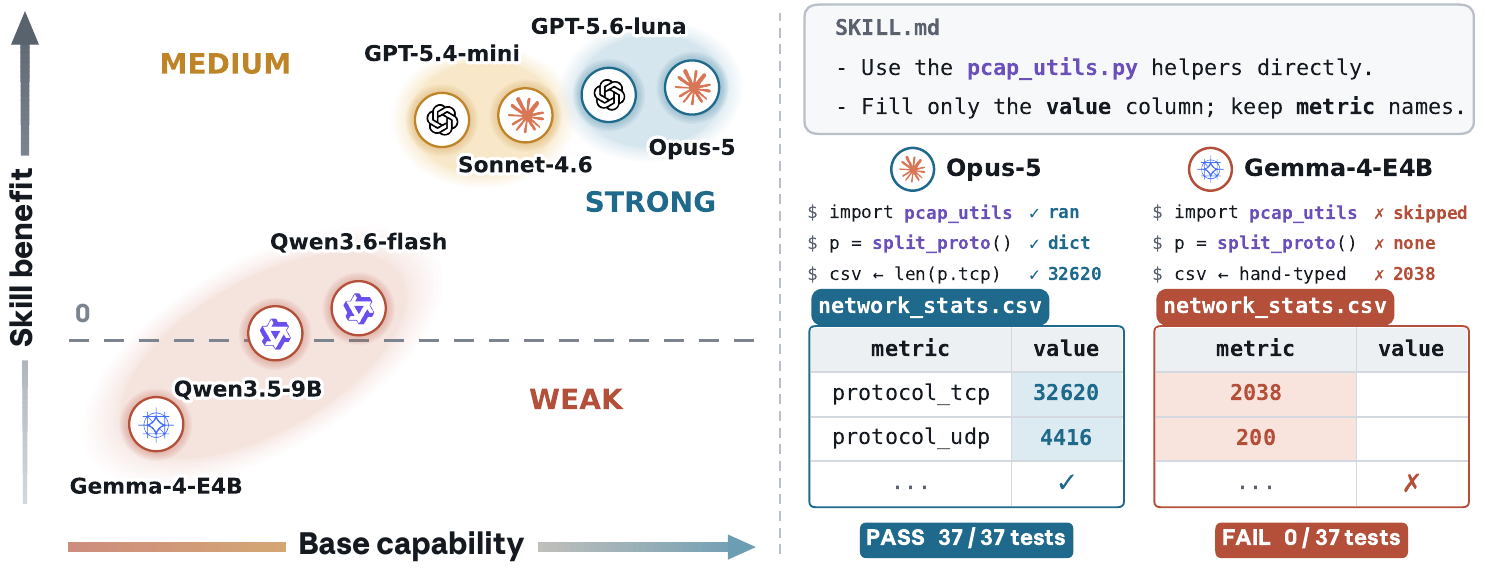}
\caption{\textbf{Skill benefit tracks base capability.}
\textbf{(a)} On $103$ skill-sensitive pairs, injected skills boost strong- and mid-tier models by $+27$\,pt on average, yet leave the weak tier nearly unchanged at $+1$\,pt.
\textbf{(b)} The gap is not missing guidance: given the same skill, the strong model follows its instructions and passes, while the weak model ignores them and fails.}
\label{fig:teaser}
\vspace{-10pt}
\end{figure*}

A common approach to harness self-evolution is to distill reusable skills from execution trajectories~\citep{wang2024agent, ouyang2026reasoningbank}.
Prior work has largely focused on efficiently extracting experiences and faithfully translating them into procedural guidance~\citep{zheng2025skillweaver, ni2026trace2skill}.
However, \textit{encoding the right experience in a skill does not guarantee that a model can use it effectively}.
We find that some skills substantially benefit stronger models while providing little benefit or even degrading performance for weaker ones. 
To characterize this disparity, we evaluate seven models with and without skills across three independently constructed benchmarks: SkillsBench~\citep{li2026skillsbench}, DocOps~\citep{jiang2026docops}, and Tessl~\citep{shaposhnikov2026framework}. We group models into capability tiers based on their performance without skills and identify $103$ tasks, where skills improve frontier models by 29.2 points on average but weak models by only 1.2 points~(\cref{fig:teaser-a}).
This limited benefit is not simply due to missing information in the skills.
Among failed weak-model runs on these pairs, 86\% violate requirements already stated in the injected skill, indicating that relevant guidance is often available but not effectively followed~(\cref{fig:teaser-b}). 
These findings suggest that harness evolution should consider not only whether an update captures useful experience, but also whether its formulation matches the executor's capabilities closely enough for that experience to be reliably translated into performance gains in downstream tasks.

To understand why weak models still fail despite receiving skill guidance, we analyze their failed execution trajectories on these tasks and identify four recurring failure modes.
First, agents struggle with \textbf{tool execution}:
they invoke unavailable tools or stop without recovering from tool errors.
Second, failures in \textbf{context management} cause agents to lose track of the current task,
driving their work toward the skill's reference material or away from the required delivery path.
Third, agents fall short in \textbf{instruction following}, leaving requirements already stated in the skill unmet.
Finally, inadequate \textbf{verification} leaves incorrect results unchecked in the final output.

Informed by this failure analysis, we propose \textit{skill adaptation}: reformulating skills to match the executor’s capabilities and facilitate harness self-evolution.
Specifically, we design a \textit{meta-skill}, a fixed, task-independent rewriting protocol that guides targeted skill revisions to eliminate the diagnosed failure patterns, enabling robust skill utilization for agentic tasks.

We first evaluate whether the \textit{meta-skill} can guide effective skill adaptation, enabling weak models to benefit from skills they previously struggled to use. In this setting, a capable rewriter receives the task, original skill, and executor’s tool list, but no execution trajectories or verifier feedback. Guided by the meta-skill, Kimi-K3~\citep{team2026kimi} produces revised skills that improve the three weak executors by 5.1--14.1 points over the original skills. With the same rewriter, the meta-skill adds 5.1--6.0 points over a generic rewriting instruction, accounting for more than half of the average gain.

\begin{wrapfigure}{r}{0.45\textwidth}
    \vspace{-12pt}
    \centering
    \includegraphics[width=0.45\textwidth]{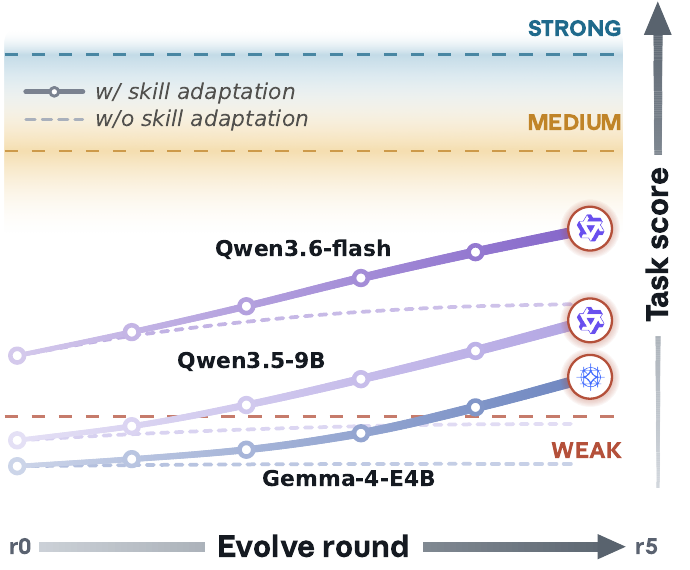}
    \caption{\textbf{Skill adaptation in self-evolution.} Over five rounds, weak executors improve steadily with skill adaptation but barely move without it. Horizontal lines: tier averages.}
    \label{fig:skill_adaptation}
    \vspace{-10pt}
    \end{wrapfigure}

We then integrate skill adaptation into a harness self-evolution loop~\citep{zhang2026darwin}. Each model alternates between executing the task and revising its own skill using the resulting trajectory and verifier feedback. After five rounds, all seven executors, from Gemma-4-E4B~\citep{gemmateam2026gemma4} to Kimi-K2.5~\citep{team2026kimi}, score 2.0--5.1 points higher with the meta-skill than under the same loop without it~(\cref{fig:skill_adaptation}). On average, both settings improve tasks associated with tool-execution and instruction-following failures, while verification-related gains are larger in self-evolution, where revisions draw on concrete failure evidence. Some failures persist even after the relevant instructions are correctly specified, revealing limits to what skill rewriting alone can achieve.
These findings suggest that harness evolution should both capture useful guidance and adapt it to the executor’s capabilities. By explicitly supporting this adaptation with the meta-skill, our method makes harness evolution more effective.

\section{Framework}
\label{sec:framework}
We first formalise the setting studied in this work.
Let $\mathcal{T} = \{T_1, \ldots, T_n\}$ be a set of skill-augmented tasks, where each task $T_i$ is paired with a human-crafted skill $S_i$.
For a model $m$, we write $s_m(T_i, S_i) \in [0, 1]$ for its score when solving $T_i$ with the skill, and $s_m(T_i, \emptyset)$ for its score without it.
We define \textbf{skill gain} as $\Delta_m(T_i) = s_m(T_i, S_i) - s_m(T_i, \emptyset)$, and our goal is to understand how $\Delta_m$ varies across models of different capabilities on the same task-skill pair.

\subsection{Skill-Task Curation}
\label{subsec:skill-task-curation}
We collect tasks together with their accompanying human-crafted skills from three benchmarks with disjoint construction protocols with details in \cref{app:bench}.
\textit{SkillsBench}~\citep{li2026skillsbench} consists of 87 skill-augmented tasks across 8 domains for STEM and engineering tasks.
\textit{DocOps}~\citep{jiang2026docops} consists of 4 types of different document operation tasks.
\textit{Tessl}~\citep{shaposhnikov2026framework} consists of 334 skill-augmented tasks with 184 different skills.
We select 7 different models from different model families.
Based on their performance without any skills on the tasks we included, we categorize them into 3 groups.
For models with \textbf{weak capabilities}, we select \textit{Gemma-4-E4B}~\citep{gemmateam2026gemma4}, \textit{Qwen3.5-9B}~\citep{qwen3.5} and \textit{Qwen3.6-flash}~\citep{qwen36plus}.
For models with \textbf{moderate capabilities}, we select \textit{GPT-5.4-mini}~\citep{openai2026gpt54mini} and \textit{Claude-Sonnet-4.6}~\citep{anthropic2026sonnet46}.
For models with \textbf{frontier capabilities}, we select \textit{GPT-5.6-luna}~\citep{openai2026gpt56sol} and \textit{Claude-Opus-5}~\citep{anthropic2026opus5}.
Due to environment issues, we run the tasks that execute reliably on our infrastructure: $87$ tasks in \textit{SkillsBench}, $186$ in \textit{DocOps} and $334$ in \textit{Tessl}, $607$ in total.

\subsection{Task Execution}
\label{subsec:protocol}
We run each task-skill pair under two conditions.
In the \textbf{w/o skill} condition, the agent receives only the task description and the task environment, yielding $s_m(T_i, \emptyset)$.
In the \textbf{w/ skill} condition, we provide the human-crafted skills for the agent, yielding $s_m(T_i, S_i)$ while keeping all other inputs and the verifier identical.
We do not evaluate skill retrieval or progressive disclosure.
Each benchmark is scored by its native judge, normalized to $[0, 1]$.
\textit{SkillsBench} ships a deterministic per-task verifier to evaluate the success rate of each task.
\textit{DocOps} opens the produced document and checks its realised state as verification for success rate.
\textit{Tessl} uses an LLM judge to grade each task against a task-completion rubric.
Because the three construction and scoring protocols are disjoint, we run every test within a benchmark and report whether the conclusion replicates across all three.

\subsection{Pair Selection}
\label{subsec:selection}
Our unit of observation is a task-skill pair $(T_i, S_i)$: the same skill appearing in two tasks counts twice, and a task injected with several skills is treated as one pair carrying the whole bundle, since the benchmarks do not permit estimating the contribution of an individual skill.
For a tier $G \in \{\mathrm{F}, \mathrm{M}, \mathrm{W}\}$ (frontier, moderate, weak), the tier gain $\Delta_G(T_i)$ is the mean of $\Delta_m(T_i)$ over the models $m \in G$.
We say that tier $G$ \textbf{benefits} from $S_i$ if
\begin{equation}
  \Delta_G(T_i) \geq \tau
  \quad\text{and}\quad
  \Delta_G(T_i) - \Delta_{\mathrm{W}}(T_i) \geq \tau,
  \qquad \tau = 0.10,
\end{equation}
i.e.\ the skill clearly helps tier $G$, and helps it clearly more than the weak tier.
A pair is \textbf{skill-sensitive} if the frontier tier benefits, or if the frontier tier is already saturated without the skill (mean score $\geq 0.95$) and the moderate tier benefits.
We discard pairs on which the weak tier already scores $\geq 0.80$ with the skill, since they leave little room for adaptation.

\begin{figure}[t]
\centering
\includegraphics[width=\textwidth]{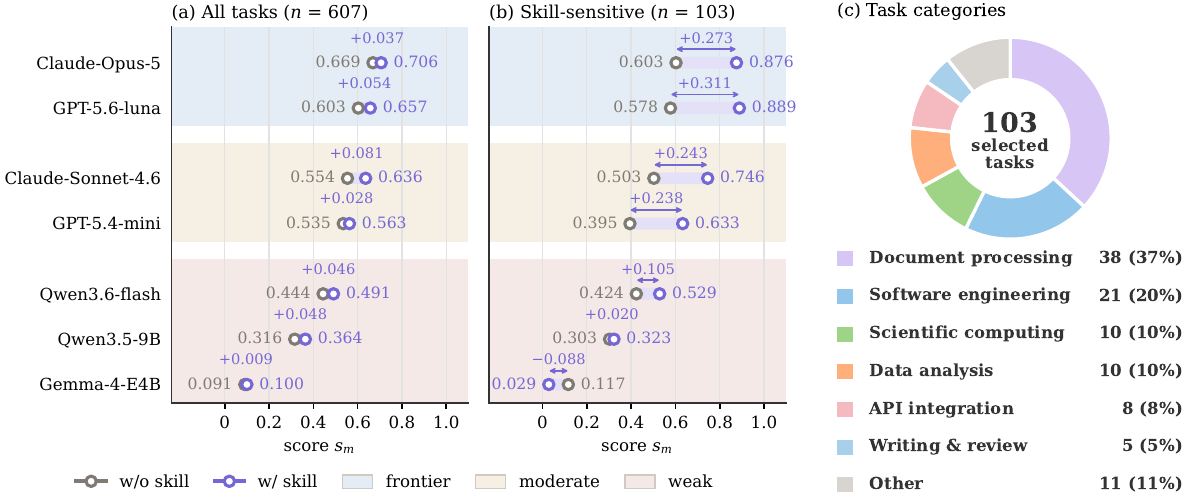}
\caption{Performance and task composition of the benchmark suite.
Bars show the score of each model w/o and w/ skill, shaded by capability tier.
\textbf{(a)}~All $607$ common-denominator tasks, averaged with equal weight over the three benchmarks.
\textbf{(b)}~The $103$ skill-sensitive pairs, averaged over pairs.
\textbf{(c)}~Domain categories of the $103$ skill-sensitive pairs.}
\label{fig:pair}
\vspace{-5pt}
\end{figure}

Averaged over all $607$ tasks, the skill yields a small gain for every model, with no clear pattern across tiers (\cref{fig:pair}).
The selection rule retains $103$ skill-sensitive pairs: $32$ from SkillsBench, $32$ from DocOps and $39$ from Tessl.
On these pairs the picture changes: the skill lifts frontier and moderate models by up to 31.1 points, yet leaves weak models nearly unchanged and even hurts the weakest one (\cref{fig:pair}).
We use these $103$ pairs as the testbed for the rest of the paper.

\section{Diagnosing Weak Executors}
To understand why weak models struggle to benefit from skills, we analyze their failures on the selected tasks (\cref{subsec:rubric}).
We then use this analysis to develop a meta-skill that guides skill adaptation to the executor's capabilities (\cref{subsec:meta-skill}).
\cref{fig:pipeline} summarizes the approach.

\subsection{Identifying Failure Modes}
\label{subsec:rubric}
For each of the $103$ skill-sensitive pairs, we read the  trajectories of the three weak models together with the injected skill under the condition of execution with skills.
For each model trajectory, we identify the first observable failure and assign its label to the task-skill pair.
A task may carry more than one label when its weak models break at different points.
Labels are based only on evidence from the execution trajectory or verifier output.

\textbf{Tool execution} covers runs that produce no deliverable. The agent stops after one or two tool calls, treats a file format such as \texttt{xlsx} as a tool, or ends with an unresolved tool error.
\textbf{Context management} covers cases where the agent treats the skill's examples or reference files as the current task. It may also save the output to the wrong path.
\textbf{Instruction following} covers cases where the agent makes multiple tool calls and produces an output, but still misses requirements already stated in the skill. It usually completes only some of these requirements.
\textbf{Verification} covers cases where the output includes all required parts, but the answer differs from the verifier's result. The execution trajectory contains no effective check for this error.

We assign $131$ labels across these $103$ tasks. Tool execution is the most common label, while Verification is the least common.
We find that $86\%$ of weak-model runs that still fail with the skill violate a requirement already stated in the injected \texttt{SKILL.md}.
This suggests that the gap between models lies in whether weak executors can act on the guidance already in the skill.
Harness evolution should therefore both \textbf{capture useful experience} and \textbf{match its formulation to the executor's capabilities}, so that the experience can translate into performance gains.

\subsection{From Failure Modes to Meta-Skill}
\label{subsec:meta-skill}
This failure analysis motivates \textbf{skill adaptation}: reformulating skills to match the executor's capabilities, thereby enabling more effective harness self-evolution.
To guide this reformulation, we design a \textbf{meta-skill}, a fixed, task-independent rewriting protocol.
It converts the four failure modes into concrete rewriting instructions, targeting the parts of a skill that weak executors struggle to follow.
We develop the meta-skill with Claude-Opus-5 as a writing assistant.

The meta-skill provides guidance for addressing the four failure modes.
For \textbf{Tool execution}, the skill should distinguish callable tools from documents and include environment checks and ways to handle likely errors.
For \textbf{Context management}, the skill should start with the user's task, the allowed scope of changes and the delivery path. It should clearly separate the current task from examples and background material, and directly include any supporting content the task needs.
For \textbf{Instruction following}, each relevant requirement should specify an \emph{action}, a \emph{target} and a \emph{completion condition}.
For \textbf{Verification}, the skill should end with an independent check: reread the result, try to recompute it using another method, compare it with an explicit threshold and repeat until the check passes.
When an execution trajectory is available, the rewriter should address its first observed failure before revising later steps.
The same meta-skill supports skill adaptation without execution evidence and self-evolution using execution trajectories and verifier feedback.
We evaluate these two uses in \cref{sec:experiments} and give the full meta-skill and rewriting protocol in \cref{app:meta,app:protocol}.


\begin{figure}[t]
\centering
\includegraphics[width=\textwidth]{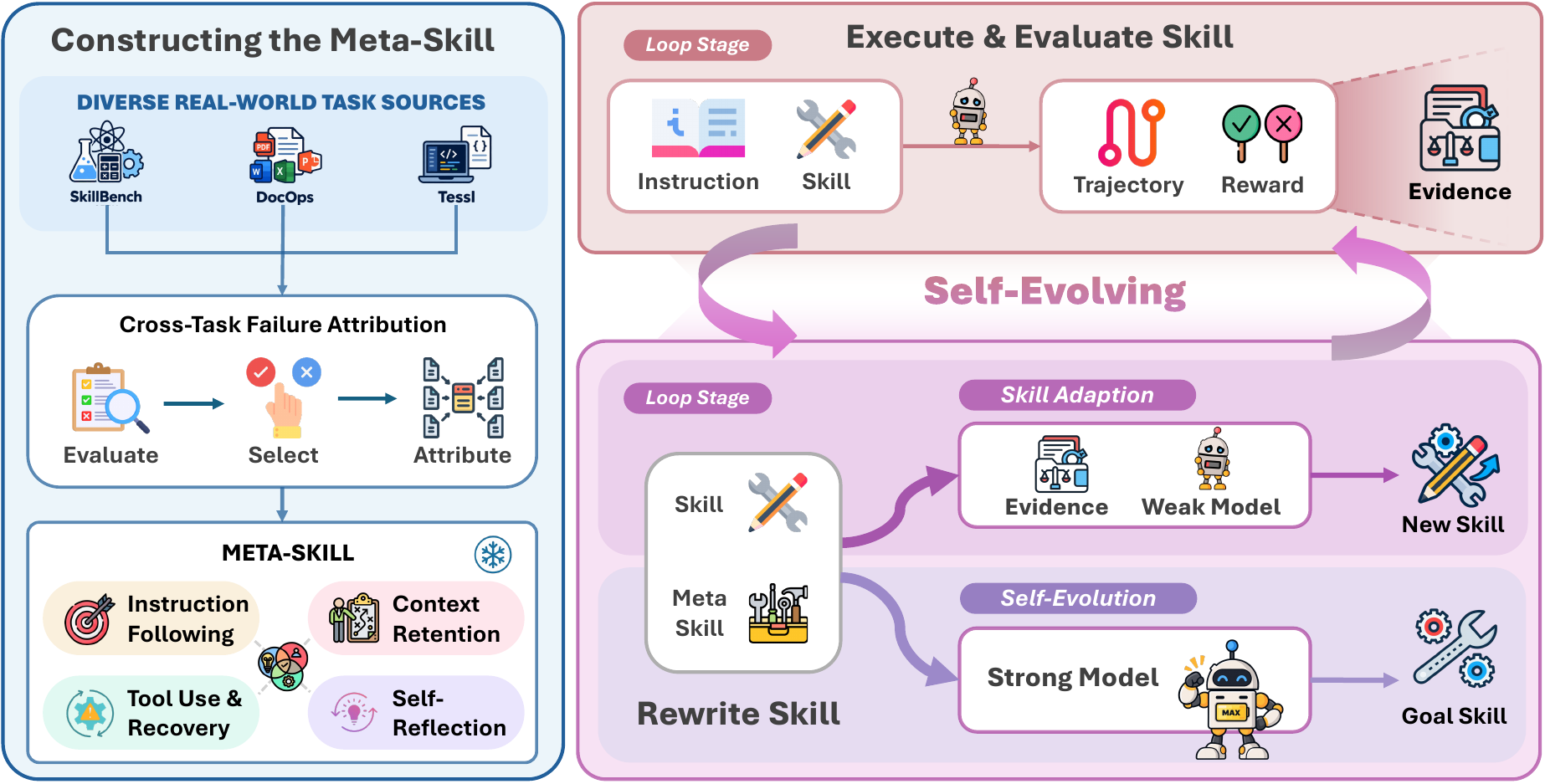}
\caption{Overview of our approach.
\textbf{Left:} the meta-skill is constructed by attributing weak-model failures across tasks from three benchmarks to four failure modes.
\textbf{Right:} the meta-skill guides a self-evolving loop of skill execution and rewriting, either in one shot by a strong model or iteratively by a weak model from its own execution evidence.}
\label{fig:pipeline}
\end{figure}
\section{Experiments}
\label{sec:experiments}
In \cref{subsec:meta-skill}, we turned the failure analysis into a meta-skill that guides skill adaptation.
We now evaluate its effectiveness on our testbed of $103$ skill-sensitive pairs in the two settings shown on the right of \cref{fig:pipeline}.
First, we ask whether the meta-skill can guide effective skill adaptation, enabling weak models to benefit from skills they previously struggled to use.
Then, inspired by recent work on agent self-improvement~\citep{li2026skillsbench,wang2026rethinking}, we evaluate the effect of integrating skill adaptation into a harness self-evolution loop.
Across both settings, we also analyse which failure modes the meta-skill repairs (\cref{subsec:analysis}).

\subsection{Skill adaptation with meta-skill}
\label{subsec:cold}
\paragraph{Setup.}
In this experiment, we evaluate whether the meta-skill can help a rewriter model to adapt a skill to weak models without checking previous execution trajectories.
We use four advanced models with strong capabilities as rewriters: GPT-5.6-sol, GPT-5.6-luna, GLM-5.3 and Kimi-K3.
We include three weak open-source models as executors: Qwen3.6-flash, Gemma-4-E4B and Qwen3.5-9B.
We run this experiment on the entire testbed of $103$ pairs.

\paragraph{Approach.}
In this experiment, the rewriter receives only the task and the skill to be revised, without any execution trajectory or feedback.
It returns either a revised skill or the original one if it decides to not revise the skill.
To keep the comparison fair, we discard revisions that leak the answer or fail to load and fall back to the original skill (\cref{app:protocol}).
Each rewriter revises every skill once, and each executor runs each revision once.
\cref{tab:rewriter} reports the mean score under each rewriting setup.

\begin{wraptable}{r}{0.46\textwidth}
\vspace{-14pt}
\centering
\caption{Skill adaptation with the meta-skill (mean scores over $103$ pairs). \emph{Original}: unrevised skills; \emph{Weak (self)}: each executor rewrites its own skill.}
\vspace{-10pt}
\label{tab:rewriter}
\footnotesize
\setlength{\tabcolsep}{2pt}
\begin{tabularx}{\linewidth}{@{}l *{4}{>{\centering\arraybackslash}X}@{}}
\toprule
Rewriter & \shortstack{Qwen\\3.6-flash} & \shortstack{Gemma\\4-E4B} & \shortstack{Qwen\\3.5-9B} & Avg. \\
\midrule
\phantom{\micon{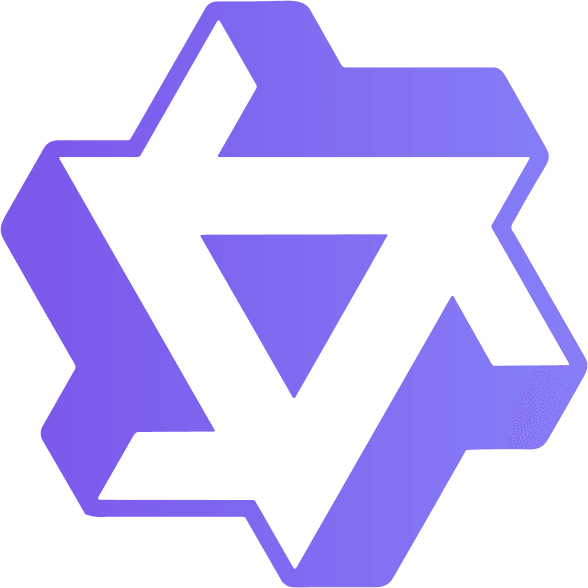}}\textit{Original} & 0.529 & 0.029 & 0.323 & 0.294 \\
\rowcolor{grouptint}\multicolumn{5}{c}{\textit{Rewritten with the meta-skill}} \\
\phantom{\micon{qwen}}Weak (self) & 0.476 & 0.056 & 0.280 & 0.271 \\
\micon{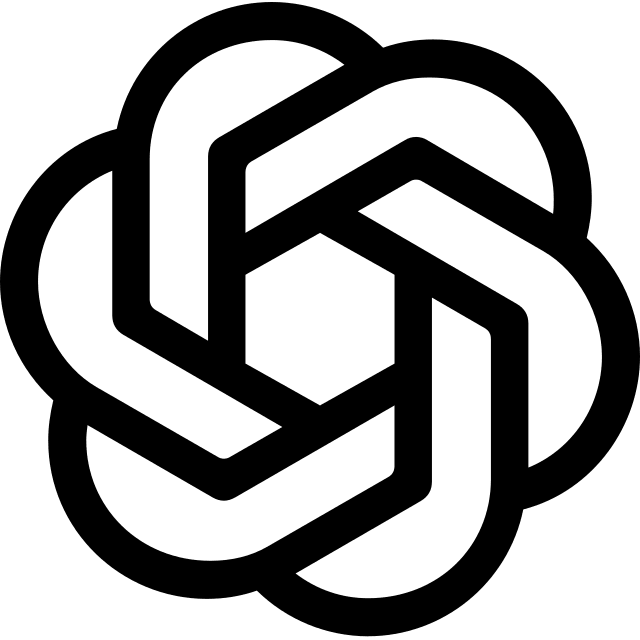}GPT-5.6-sol & 0.563 & 0.139 & 0.303 & 0.335 \\
\micon{openai}GPT-5.6-luna & 0.602 & 0.119 & 0.307 & 0.343 \\
\micon{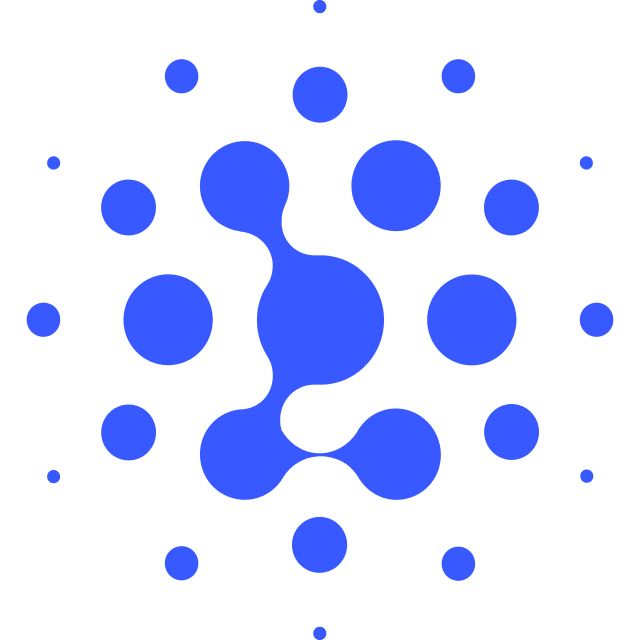}GLM-5.3 & \textbf{0.626} & 0.138 & \textbf{0.374} & 0.379 \\
\rowcolor{besttint}\micon{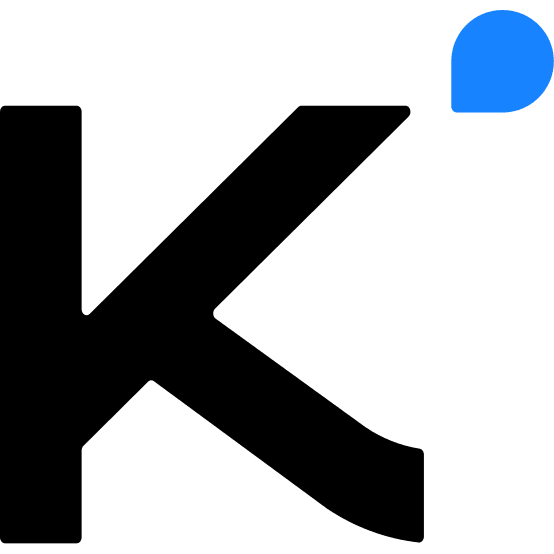}Kimi-K3 & 0.616 & \textbf{0.170} & \textbf{0.374} & \textbf{0.387} \\
\bottomrule
\end{tabularx}

\end{wraptable}

\paragraph{Results.}
With a capable rewriter, the meta-skill turns skills that weak models could not use into skills they can: all four strong rewriters raise the average, and Kimi-K3 and GLM-5.3 improve all three executors, lifting the average by about $+0.09$.
The weak executors rewriting their own skills, by contrast, end up below the original (\cref{tab:rewriter}).
The weakest executor gains the most: under Kimi-K3, Gemma-4-E4B rises from $0.029$ to $0.170$.
This suggests that weak models can benefit more from human-crafted skills through skill adaptation by a capable rewriter.
\WFclear

\begin{takeaway}
Through skill adaptation, a capable rewriter turns human-crafted skills that weak models could not use into skills they can use effectively.
\end{takeaway}

\begin{wrapfigure}{r}{0.46\textwidth}
\vspace{-14pt}
\centering
\includegraphics[width=0.44\textwidth]{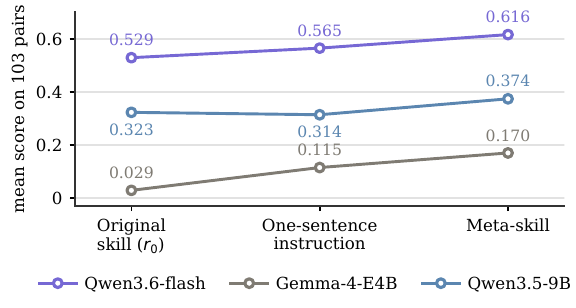}
\caption{Skill adaptation by Kimi-K3 with the meta-skill or a one-sentence instruction; each weak executor runs the revised skill once on all $103$ pairs.}
\label{fig:no-meta}
\vspace{-10pt}
\end{wrapfigure}
\paragraph{Skill rewriting without meta-skill.}
Since the best rewriters above are strong models, the gains may owe as much to the rewriter as to the meta-skill.
To separate the two, we replace the meta-skill with a one-sentence instruction that asks the rewriter to revise the skill so that it works for all models, and keep everything else unchanged (\cref{app:meta}).
We use Kimi-K3, the strongest rewriter in \cref{tab:rewriter}, under the same setup as above (\cref{fig:no-meta}).
We find that a capable rewriter does help on its own: with the one-sentence instruction alone, Kimi-K3 improves two of the three executors.
The gain is uneven, however, and Qwen3.5-9B ends slightly below the original skill. 
With the meta-skill, the same rewriter improves all three executors over the one-sentence instruction, and the added gain is consistent across them.
This suggests that guidance on weak executors' failure modes helps a capable rewriter produce more effective skill revisions.

\begin{takeaway}
Distilling executor failure modes into a meta-skill improves skill adaptation by providing explicit guidance on what to revise and how.
\end{takeaway}

\subsection{Self-evolution with meta-skill}

\label{subsec:rsi}
So far, we have discussed whether the meta-skill can guide effective skill adaptation without access to failure trajectories.
We now integrate skill adaptation into a more realistic harness self-evolution loop~\citep{chen2026recursive}: each model alternates between executing the task and revising its own skill using the resulting trajectory and verifier feedback.

\paragraph{Setup.}
We now let each weak model improve its own skill, so the rewriter and the executor are the same model.
Unlike one-shot rewriting, the rewriter now sees the evidence of failure: in round $t$, it receives the task, the skill from round $t-1$, the executor's tool list, and the trajectory and verifier output of round $t-1$.
It returns a revised skill, which the executor then runs on the task, and we apply the same validity checks as in one-shot rewriting.
We repeat this loop for five rounds, with each round revising the skill produced by the previous one.
To illustrate the effectiveness of meta-skill in the practice of self-evolution, we compare with a baseline that does not include meta-skill in the system prompt context.
We use the same three weak executors as above and run every round once on all 103 pairs.
Since this setting spans several rounds, we measure gains against a baseline run in the same time window, in which each executor reruns the original skill without revision.
To test whether the effect extends beyond the weakest models, we also run four additional executors: Qwen3.5-27B, Gemma-4-26B, Qwen3.5-plus and Kimi-K2.5.

\begin{figure}[t]
\centering
\includegraphics[width=\textwidth]{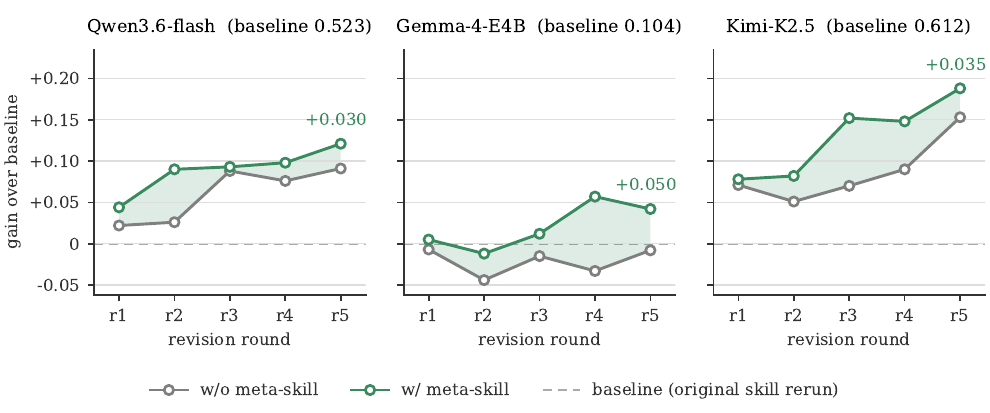}
\caption{Self-evolution with and without the meta-skill. Each executor rewrites its
own skill from the previous round's trajectory and verifier output over five rounds, and runs every
round once on all $103$ pairs. Curves show the gain over the baseline that reruns the original skill;
annotations mark the gap between the two arms at round~5. Results for the remaining four executors are in \cref{fig:rsi-appendix}.}
\label{fig:rsi}
\end{figure}

\paragraph{Results.}
\cref{fig:rsi} reports the gain of each round over this baseline for Qwen3.6-flash, Gemma-4-E4B and Kimi-K2.5. \cref{fig:rsi-appendix} shows the remaining four executors, and \cref{tab:rsi-full} in \cref{app:rsi} reports every round for all seven executors.
With the meta-skill, every executor ends round 5 above the arm without it (\cref{fig:rsi,fig:rsi-appendix}).
We observe steady performance improvement across different executors, suggesting the effectiveness of meta-skill during the self-evolution of agents.
Without the meta-skill, self-revision helps little and is unstable: Qwen3.5-9B peaks at round 3 and then falls back almost to the baseline, and Gemma-4-E4B, the weakest executor, stays below its baseline in every round.
The meta-skill does not make the curves monotonic, since Qwen3.5-9B still drops after round 3 and Gemma-4-E4B dips at round 2, but both end higher, and Gemma-4-E4B stays above its baseline from round 3 on.
The four larger executors move in the same direction, with round-5 gaps of $+0.046$ on Qwen3.5-27B, $+0.020$ on Gemma-4-26B, $+0.025$ on Qwen3.5-plus, and $+0.035$ on Kimi-K2.5 (absolute scores in \cref{tab:rsi-full}).
This is the opposite of what we saw in one-shot rewriting, where rewriting with the meta-skill hurt two of the three weak executors.
The difference is the failure evidence: once the rewriter can see how the skill failed, even a weak model can make use of the meta-skill.
Because each round is run once, we do not read much into the exact size of these $0.02$--$0.05$ gaps; what we rely on is that their sign is the same for all seven executors.

\begin{takeaway}
Given evidence of failure, weak models can use the meta-skill to self-evolve better skills, ending above self-evolution without it.
\end{takeaway}
\begin{table}[t]
\centering
\caption{Score gains from the meta-skill over rewriting without it, averaged over pairs with each failure label. \emph{Skill adaptation} uses Kimi-K3 rewrites; \emph{Self-evolution} uses round-5 skills. $n$ counts pairs, and labels can overlap. Shading indicates the size of the gain.}
\label{tab:rubric}
\small
\setlength{\tabcolsep}{4pt}
\renewcommand{\arraystretch}{1.15}
\begin{tabular}{l r ccc c ccc c}
\toprule
\multirow{2}{*}{Failure mode} & \multirow{2}{*}{$n$} & \multicolumn{4}{c}{Skill adaptation} & \multicolumn{4}{c}{Self-evolution} \\
\cmidrule(lr){3-6}\cmidrule(lr){7-10}
 & & \twoline{\micon{qwen}Qwen}{3.6-flash} & \twoline{\micon{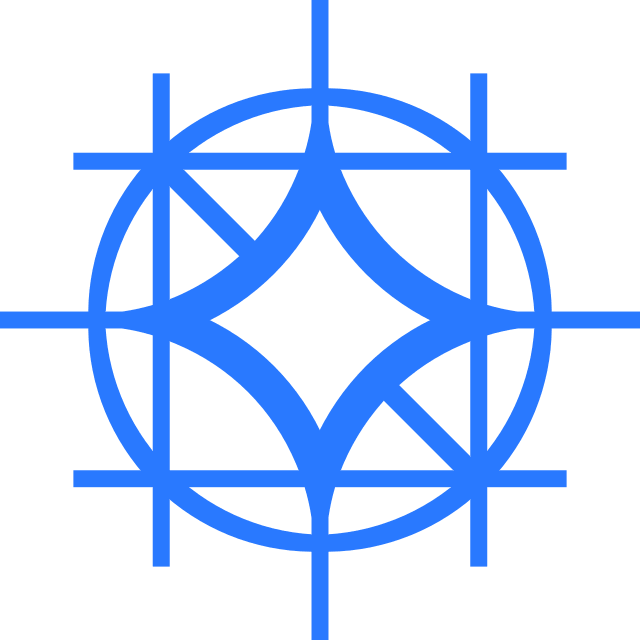}Gemma}{4-E4B} & \twoline{\micon{qwen}Qwen}{3.5-9B} & Avg. & \twoline{\micon{qwen}Qwen}{3.6-flash} & \twoline{\micon{gemma}Gemma}{4-E4B} & \twoline{\micon{qwen}Qwen}{3.5-9B} & Avg. \\
\midrule
Execution & 51 & \cellcolor{takeawayblue!11}+0.067 & \cellcolor{takeawayblue!5}+0.033 & \cellcolor{takeawayblue!11}+0.067 & \cellcolor{takeawayblue!9}\textbf{+0.056} & \cellcolor{takeawayblue!7}+0.043 & \cellcolor{takeawayblue!0}+0.004 & \cellcolor{takeawayblue!24}+0.149 & \cellcolor{takeawayblue!10}\textbf{+0.065} \\
Instruction & 35 & \cellcolor{takeawayblue!4}+0.027 & \cellcolor{takeawayblue!10}+0.065 & \cellcolor{takeawayblue!12}+0.073 & \cellcolor{takeawayblue!9}\textbf{+0.055} & \cellcolor{lossink!11}$-$0.069 & \cellcolor{takeawayblue!4}+0.026 & \cellcolor{takeawayblue!30}+0.182 & \cellcolor{takeawayblue!7}\textbf{+0.046} \\
Context & 28 & \cellcolor{lossink!6}$-$0.041 & \cellcolor{takeawayblue!23}+0.141 & \cellcolor{takeawayblue!34}+0.209 & \cellcolor{takeawayblue!17}\textbf{+0.103} & \cellcolor{takeawayblue!2}+0.015 & \cellcolor{takeawayblue!19}+0.114 & \cellcolor{lossink!17}$-$0.107 & \cellcolor{takeawayblue!1}\textbf{+0.007} \\
Verification & 17 & \cellcolor{takeawayblue!21}+0.127 & \cellcolor{takeawayblue!3}+0.018 & \cellcolor{lossink!22}$-$0.136 & \cellcolor{takeawayblue!0}\textbf{+0.003} & \cellcolor{takeawayblue!39}+0.234 & \cellcolor{takeawayblue!3}+0.018 & \cellcolor{takeawayblue!9}+0.059 & \cellcolor{takeawayblue!17}\textbf{+0.104} \\
\bottomrule
\end{tabular}
\end{table}

\subsection{Analysis}
\label{subsec:analysis}
To compare what skill adaptation improves in the two settings, we group the $103$ pairs by their failure labels from \cref{subsec:rubric} and compare scores with and without the meta-skill within each group (\cref{tab:rubric}).


\paragraph{Skill adaptation addresses problems visible in the text.}
We can tell from the skill text whether it names callable tools, gives each requirement a completion condition, and states the task and delivery path at the start.
In \cref{fig:case}, the rewrite specifies the delivery target and required \texttt{EPSG:4087} projection,
requires ISO~8601 timestamps, and adds an independent result check.
These changes make explicit the steps the executor would otherwise need to infer from reference material.
In one-shot skill adaptation, the meta-skill improves performance on Execution, Instruction, and Context, with mean score gains of $+0.056$, $+0.055$, and $+0.103$, respectively (\cref{tab:rubric}).

\paragraph{Self-evolution addresses problems revealed during execution.}
For Verification, the skill text can specify a check, but execution trajectories reveal whether the executor actually performs it.
Self-evolution uses this evidence to guide subsequent skill revisions.
In \cref{fig:case}, the executor without the meta-skill keeps reporting the same incorrect distance across all five rounds, despite instructions to rerun the script on errors.
With the meta-skill, the revised skill requires rereading \texttt{answer.json} and recomputing the distance, and Qwen3.6-flash passes four of five rounds.
The mean score gain from the meta-skill on Verification is $0.101$ larger in self-evolution than in one-shot skill adaptation (\cref{tab:rubric}).
The meta-skill also improves scores on Execution and Instruction in self-evolution, where trajectories expose tool errors and missed requirements that can guide further revisions.

\paragraph{What the meta-skill cannot repair is the executor.}
The failures that remain are the executor's own.
Behind the negative cells of \cref{tab:rubric}, the revised skill states the right procedure and the executor does not carry it out: it edits a script without rerunning it, repeats a failed call, stops before writing the artefact, or, when a task has several constraints, keeps the one the skill highlights and drops the rest.
\cref{fig:case} is typical: with the meta-skill the skill covers all four dimensions after one round and Qwen3.6-flash passes four of five; in the failing round it fixes the script but never reruns it.
A skill can tell a weak executor what to do; getting it done depends on the model. \cref{app:cases} gives more cases.

\begin{figure}[t]
\centering
\begin{tcbraster}[raster columns=2, raster equal height, raster column skip=6pt, raster valign=top]
\begin{casebox}{\casetitle{w/o meta-skill}{skill after round 1}}
\begin{plainbox}{\badmark}\caseh{Task Scope} Never stated: six sections of reference material, none says what to deliver.\end{plainbox}
\begin{plainbox}{\badmark}\caseh{Antimeridian \& Projection} Shift longitudes to $[0, 360]$; project with a custom \texttt{+proj=aeqd} CRS.\end{plainbox}
\begin{plainbox}{\badmark}\caseh{Format \& Save} Write \texttt{answer.json} and print ``Done''; the timestamp format is never checked.\end{plainbox}
\begin{plainbox}{\badmark}\caseh{Execution Protocol} Run the script and rerun on error; nothing checks the reported value.\end{plainbox}
\tcblower
\casereward{\badmark}{\shown{\badmark}\,\badmark\,\badmark\,\badmark\,\badmark}{Wrong in every round: the skill stays a list of reference notes with no delivery target and no result check, so the same wrong distance survives every rewrite.}
\end{casebox}
\begin{metacasebox}{\casetitle{w/ meta-skill}{skill after round 1}}
\begin{metabox}{Context management}{\okmark}\caseh{Task Scope \& Delivery} State the inputs, the output path and its keys; the rest is reference material.\end{metabox}
\begin{metabox}{Tool execution}{\okmark}\caseh{Projection Rule} Use \texttt{EPSG:4087} only; do not invent a custom projection.\end{metabox}
\begin{metabox}{Instruction following}{\okmark}\caseh{Format Output} Convert \texttt{time} to ISO 8601 and check the format before writing.\end{metabox}
\begin{metabox}{Verification}{\okmark}\caseh{Verification Loop} Re-read \texttt{answer.json}, recompute the distance, loop until it matches.\end{metabox}
\tcblower
\casereward{\badmark}{\shown{\okmark}\,\okmark\,\badmark\,\okmark\,\okmark}{Right from round 1: scope, projection, format and read-back provide guidance for all four failure modes. The one miss is a script edited but not rerun, which no skill text can catch.}
\end{metacasebox}
\end{tcbraster}
\caption{Skills after the first round of self-evolution by Qwen3.6-flash on one task, with and without the meta-skill. Rows are aligned by the failure mode addressed by the displayed guidance. Marks give the task reward from round 0, the original skill, to round 5; the framed mark is the round shown.}
\vspace{-20pt}
\label{fig:case}
\end{figure}

\section{Related Work}

\paragraph{Agent Skills.}
Agent skills are composable packages of instructions, code, and resources that agents load on demand, extending model capabilities without changing its parameters~\citep{anthropic2025skills,xu2026agent,jiang2026sok,weng2026harness}. 
They have become a standard layer of the agent harness alongside tools and memory~\citep{yang2024swe, weng2026harness}, and are populated in two ways: distilled from the trajectories of capable models into reusable programs or workflows~\citep{wang2023voyager, wang2024agent, zhang2026coevoskills}, or written by human experts and distributed through public registries~\citep{gao2026registry,liu2026agent}. 
A growing set of benchmarks asks whether injected skills actually help, covering STEM and software-engineering tasks~\citep{li2026skillsbench,han2026swe}, verifiable document operations~\citep{jiang2026docops}, skill evaluation at scale~\citep{shaposhnikov2026framework}, and realistic retrieval settings~\citep{liu2026agent}. 
The ecosystem is expanding fast, with tens of thousands of skills on public registries~\citep{ling2026agent} and active work on retrieval~\citep{zheng2026skillrouter}, compression~\citep{gao2026skillreducer}, and security~\citep{liu2026skillsvote}. 
Throughout, a skill is written once and shipped unchanged across models. Early evidence suggests this is fragile~\citep{yu2026skill}, but how a fixed skill's benefit varies with model capability remains unexamined.

\paragraph{Harness Evolution.}
Harness evolution lets an agent improve itself by rewriting its own harness from execution feedback~\citep{zhang2026darwin, lee2026meta, zhang2026hyperagents}.
Early systems search over agent code or prompts~\citep{hu2025automated,zhang2026darwin, zhang2026agentic}, and recent work makes the loop recursive, evolving the improvement procedure itself~\citep{zhang2026hyperagents,lee2026recursive,wang2026metaskill}.
Two designs are closest to ours. Meta-Harness~\citep{lee2026recursive} uses a stronger external optimiser to tune the harness of a weaker model, mirroring our skill adaptation setting; Self-Harness~\citep{zhang2026self} lets a model mine its own failure traces and propose bounded edits, mirroring our self-evolution setting. 
At the skill level, revisions are driven by failure traces~\citep{liu2026skillrevise,alzubi2026evoskill}, verifier co-evolution~\citep{zhang2026coevoskills}, or held-out gating~\citep{yang2026skillopt}.
Yet we differ from these works in two ways.
First, we focus on the problem of strong-to-weak model mismatch, where a skill is well-suited for a strong model but executed poorly by a weak model.
Second, we examine whether weak models can improve their own skills, resonating with recent work on recursive harness improvement~\citep{chen2026recursive}.
\section{Conclusion}
\label{sec:conclusion}
We showed that the benefit of a harness is not a property of the harness alone: the same skill that lifts frontier models leaves weak models unchanged or worse, even though the requirements they violate are already written in it.
We traced the loss to four failure modes in how a skill is written, and distilled them into a meta-skill, a fixed protocol that lets a capable rewriter adapt skills without execution evidence and lets weak models improve their own skills over rounds of self-evolution.
For harness self-evolution, this suggests two practical guidelines: the benefit of a harness should be evaluated per capability tier rather than with a single representative model, and the evolution loop should be given diagnostic guidance on how a weak executor fails, since evidence alone leads weak rewriters to drift while the meta-skill lets them hold their gains.

\paragraph{Limitations.}
We evaluate skills already supplied in the executor's context, leaving retrieval and progressive disclosure outside our scope.
Multiple skills are evaluated as a bundle, and self-evolution uses a fixed budget of five rounds without updating the underlying model parameters.

\paragraph{Discussion.}
Our results suggest that harness self-evolution benefits from explicit guidance on turning execution feedback into revisions the executor can use.
The meta-skill provides this guidance for skill adaptation by connecting recurring failure modes to concrete rewriting instructions, helping weak executors improve their own skills from execution evidence.
A useful next step is to track whether each revision resolves the failure that motivated it, alongside changes in task performance.
This could guide later updates toward persistent problems and help retain revisions that remain useful across repeated attempts.
Similar guidance could also be explored for tool interfaces and agent workflows, extending the role of failure analysis in harness self-evolution.

\section*{AI use statement}
We used generative AI tools to assist with coding and with drafting, revising, and polishing the manuscript.
The authors reviewed AI-assisted code and text, verified the scientific claims and citations, and remain responsible for the research design, failure analysis, and interpretation of results.
No human-subject data were collected.
We take full responsibility for the manuscript, reported results, and accompanying artifacts, including AI-assisted content.

\section*{Reproducibility Statement}
We describe the benchmark suite and task selection in Section~\ref{sec:framework} and Appendix~\ref{app:bench}. The rewriting protocol, validity checks, and scoring conventions are detailed in
Appendix~\ref{app:protocol}, and the full meta-skill and baseline prompts are provided in Appendix~\ref{app:meta}. After the review process concludes, we will open-source the code for evaluation, skill adaptation, and self-evolution, together with the task lists and experiment configurations, to support reproduction of our results and further research on harness self-evolution.
\bibliography{iclr2027_conference}
\bibliographystyle{iclr2027_conference}

\clearpage
\appendix
\crefalias{section}{appendix}       
\crefalias{subsection}{appendix}
\newtcolorbox{promptbox}[2][]{enhanced, breakable,
  colback=grouptint, colframe=casegray, boxrule=0.5pt, arc=1.5pt,
  left=6pt, right=6pt, top=4pt, bottom=4pt,
  fonttitle=\bfseries\small, fontupper=\small, coltitle=black, colbacktitle=casegray!15,
  before upper={\setlength{\parskip}{4pt}}, title={#2}, #1}

\newtcolorbox{algshade}[1]{enhanced, colback=#1, colframe=#1, boxrule=0pt, arc=1.5pt,
  left=0pt, right=2pt, top=1pt, bottom=1pt, boxsep=0pt, before skip=3pt, after skip=1pt}

\section{Benchmarks and Task Suite}
\label{app:bench}
\cref{tab:bench} summarises the three benchmarks from which the $607$ tasks and $103$ skill-sensitive pairs are drawn.
We chose them to cover different skill authors, agent harnesses and judges, so that every conclusion in the paper can be tested on three independently built suites and is reported only when it replicates on all of them, not on a single construction protocol.

\begin{table}[h]
\centering
\caption{The three benchmarks. \emph{Tasks} counts the tasks that run reliably in our environment; \emph{Pairs} counts the skill-sensitive pairs selected in \cref{subsec:selection}.}
\label{tab:bench}
\small
\setlength{\tabcolsep}{3pt}
\begin{tabular}{l l c c l l}
\toprule
Benchmark & Domain & Tasks & Pairs & Executor harness & Judge \\
\midrule
SkillsBench & STEM and engineering, 8 domains & 87 & 32 & OpenHands & deterministic tests \\
DocOps & Excel, Word, PowerPoint, PDF & 186 & 32 & Claude Code & document-state checks \\
Tessl & developer tooling and APIs & 334 & 39 & Claude Code & LLM judge with rubric \\
\bottomrule
\end{tabular}
\end{table}

\paragraph{SkillsBench~\citep{li2026skillsbench}.}
Each task is a self-contained engineering or scientific problem, e.g.\ computing the earthquake farthest from a plate boundary, tuning an MPC controller, solving a PDDL planning instance or dispatching a power grid, and ships with one or more human-written skills (a \texttt{SKILL.md} with optional helper scripts) that a domain expert would consult for the task.
The executor runs inside the OpenHands harness and its output is scored by a per-task deterministic verifier whose assertions never read the skill body.
The $32$ selected pairs span the full range of domains, from control and planning to bioinformatics and finance.

\paragraph{DocOps~\citep{jiang2026docops}.}
Tasks edit or produce office documents of four types and are organised in tiers of increasing scope: atomic edits of one document, composite operations on one document type, cross-document operations, single-document workflows, and a set of harder publication and repair tasks.
Every task injects the skill of its document type (\texttt{xlsx}, \texttt{docx}, \texttt{pptx} or \texttt{pdf}), a long reference document with helper scripts such as a LibreOffice recalculation script.
The executor runs inside Claude Code; the checker opens the produced document and verifies its realised state, e.g.\ cell formulas, deleted text spans or theme colours.
Of the $32$ selected pairs, $10$ are atomic, $10$ composite, $5$ cross-document, $1$ workflow and $6$ publication or repair tasks.

\paragraph{Tessl~\citep{shaposhnikov2026framework}.}
Tasks pair a realistic developer request, e.g.\ setting up a payment project, configuring tracing or reviewing a cookbook, with a skill published by the tool's vendor or open-source maintainers, including Stripe, Vercel, ClickHouse, Sentry, Shopify and Firecrawl.
The executor runs inside Claude Code, and because these tasks have no single correct artefact, an LLM judge grades the result against a task-specific rubric; we use its normalised score directly.
All $39$ selected pairs are single-skill tasks, each shipping exactly one vendor skill.

\section{Rewriting Protocol}
\label{app:protocol}
Both experiments in \cref{sec:experiments} share one protocol, which constrains only the interface of the rewriter and the validity of its output, never the content of the rewrite.
\cref{alg:evolve} states it: a single rewrite step, \textsc{Adapt}, is called once without execution evidence in skill adaptation, and five times in self-evolution, each time with the score, trajectory and verifier output of the previous round.

\begin{algorithm}[h]
\caption{Skill adaptation. \textsc{Adapt} is the shared rewrite step; the two settings differ only in the rewriter $A$, the evidence $E$ and the number of rounds.}
\label{alg:evolve}
\begin{algorithmic}[1]
\Require task $T$ with instruction $I$ and verifier $V$; original skill $S_0$; executor $M$ with tool interface $\mathcal{U}_M$; rewriter $A$ with system prompt $P$; validity gate $G$
\Statex
\Procedure{Adapt}{$S, E$} \Comment{one rewrite of skill $S$ given evidence $E$}
  \State $o \gets A\big(P;\; I,\ \mathcal{U}_M,\ S,\ E\big)$ \Comment{a revised skill or \texttt{NO\_CHANGE}}
  \If{$o$ is not parseable} retry once with the parse error appended \EndIf
  \If{$o = \texttt{NO\_CHANGE}$ \textbf{or} $G(o, S)$ rejects $o$} \Comment{gate: answer leakage, loadability}
     \State \Return $S$
  \EndIf
  \State \Return $o$
\EndProcedure
\algstore{evolve}
\end{algorithmic}
\begin{algshade}{coldtint}
\begin{algorithmic}[1]
\algrestore{evolve}
\State \textbf{Skill adaptation} \Comment{strong rewriter $A$, no execution evidence}
\State \quad $S_1 \gets \textsc{Adapt}(S_0, \emptyset)$
\State \quad $(\tau_1, s_1) \gets \textsc{Run}(M, T, S_1)$;\; \Return $s_1$ \Comment{one rollout, scored by $V$}
\algstore{evolve}
\end{algorithmic}
\end{algshade}
\begin{algshade}{evotint}
\begin{algorithmic}[1]
\algrestore{evolve}
\State \textbf{Self-evolution} \Comment{the executor rewrites its own skill, $A = M$}
\State \quad $(\tau_0, s_0) \gets \textsc{Run}(M, T, S_0)$
\For{$t = 1, \ldots, 5$}
  \State $E_{t-1} \gets (s_{t-1}, \tau_{t-1}, V(\tau_{t-1}))$ \Comment{score, trajectory and verifier output of the last round}
  \State $S_t \gets \textsc{Adapt}(S_{t-1}, E_{t-1})$
  \State $(\tau_t, s_t) \gets \textsc{Run}(M, T, S_t)$
\EndFor
\State \Return $s_1, \ldots, s_5$
\end{algorithmic}
\end{algshade}
\end{algorithm}

\paragraph{Inputs to the rewriter.}
The two arms of every experiment receive an identical user prompt and differ only in the system prompt (\cref{app:meta}).
The user prompt contains, in order:
(i)~the task instruction, declared authoritative for the task, the allowed scope of changes and the delivery path;
(ii)~the executor interface, i.e.\ the complete tool list and parameter schema of OpenHands for SkillsBench, or the tool names exposed by Claude Code for DocOps and Tessl, with the note that skill names, file names and section headings are not callable tools;
(iii)~the current skill, every \texttt{SKILL.md} verbatim, together with the list of auxiliary files in the skill folder;
(iv)~in the self-evolution setting only, the result of the last attempt (\textsc{passed}, \textsc{failed} or \textsc{partial} with its score), a step-by-step digest of the executor's trajectory, and the verifier output;
(v)~a one-paragraph job statement.
In the self-evolution setting the job statement reads ``You know what the weak model did, which checks failed, and what the test output said. Rewrite the skill so that the same weak model does better on the next attempt. Do not encode expected values or test-specific answers.''; in the skill adaptation setting, where no execution has taken place, the first sentence is replaced by ``Your evidence is the task instruction, the skill as it stands, and the executor's tool interface.''

\paragraph{Output contract.}
The rewriter returns the complete new contents of every file it changes, one block per file, delimited by \texttt{<<<FILE path=...>>>} and \texttt{<<<END>>>}; files it does not return are kept unchanged.
It may restructure freely: reorder, merge, split or delete what the executor cannot act on.
Alternatively it returns a single \texttt{<<<NO\_CHANGE>>>} block with a one-line reason and a list of the missing evidence, in which case the executor reruns the unchanged skill.
There is no patch format, no required document template and no diagnosis schema; whether to change the skill at all, how much, and in what form is left entirely to the rewriter.

\paragraph{Validity checks.}
The harness applies two checks to each returned file.
\emph{Loadability}: the file must be non-empty, its path must name an existing skill file, its YAML frontmatter must survive with the \texttt{name} field unchanged (the harness registers skills by name), and it must not exceed a $64{,}000$-character budget, since the skill is injected whole into the executor's context.
\emph{Answer leakage}: we collect every specific token (identifiers of at least eight characters containing a digit or underscore, and decimals with at least two places) that occurs in the hidden tests or reference solution but neither in the public task material, including the original skill, nor in the evidence shown to the rewriter.
A revision that introduces one such expected value, or two such identifiers, is rejected; a token the rewriter could have read in the verifier output is counted but not blocked, since that route is the design of the evidence arm.
A response that cannot be parsed is sent back once with the parse error; a loadability or leakage rejection is terminal.
When a revision is rejected the executor runs the unchanged skill and the cell is still scored, so invalid rewrites lower the mean of their arm rather than disappearing from it; a \texttt{NO\_CHANGE} response is scored in the same way.

\paragraph{Scoring conventions.}
Every model--task--condition cell is run once.
Skill adaptation reports the executor's mean score on the $103$ pairs under each rewriter.
Self-evolution forms a linear chain of five rounds with no selection of the best round and no rollback, and reports each round against a baseline in which the same executor reruns the original skill in the same time window, so that drift of the serving endpoints over the experiment does not enter the gain.

\section{The Meta-Skill and the Baseline Prompts}
\label{app:meta}
The meta-skill is a fixed, task-independent document placed in the rewriter's system prompt; its full text is given below.
It is followed by the two system prompts it is compared against: the no-meta baseline of the self-evolution experiment, which keeps the role description and the two hard rules but gives no rewriting guidance, and the one-sentence instruction used in \cref{fig:no-meta}.

\begin{promptbox}{Meta-skill (system prompt of the \emph{w/ meta-skill} arm)}
\textbf{Skill Adaptation}

You are editing one \textbf{skill} --- a document that is injected into a coding agent's context before it starts a task. The agent that will read your output is a small, weak model. It has already tried this task once with the current version of the skill. You are given what happened. Produce a better version.

The skill is the only thing you may change. You cannot change the task, the agent, or the tests.

\textbf{First locate the break, then fix that}

Across 602 measured model-task cells, weak-model failure sorts into four classes by how far the agent got before it stopped. Read the trajectory and decide which of the four below is the \textbf{earliest} thing that went wrong; a deeper fix has no effect while a shallower break remains. Apply the prescription for that class --- edit the passages it implicates, and leave the rest of the document alone.

\begin{enumerate}[nosep, leftmargin=1.4em]
\item \textbf{EXE --- no action, or errors it never recovered from.} One or two tool calls, no file written; it read a format name (\texttt{xlsx}, \texttt{pdf}) as if it were a callable tool, or hit an error and stopped.
\emph{Prescription:} near the top, say which names are callable and which are documents, add an environment preflight, and give a fallback for the first likely failure. Give commands the agent can run verbatim.
\item \textbf{CTX --- the skill displaced the task.} It succeeded \emph{without} the skill and failed \emph{with} it: work went into the skill's own examples, files it was only asked to patch, or an artefact outside the required path.
\emph{Prescription:} state the user's task, the allowed modification scope, and the delivery path early, and say in one line that the rest of the document is reference material rather than a new request. If a step needs content from an auxiliary file, inline that content.
\item \textbf{IF --- requirements not met.} Files written, plenty of tool calls, and the failed check is a requirement the text already states. It did part of a multi-part requirement and stopped.
\emph{Prescription:} rewrite each implicated requirement as one line of \textbf{action --- target --- completion condition}, including the negative branch (``if X already exists, stop and change nothing''). Writing more \emph{about} the requirement does not help; the text was already there.
\item \textbf{REF --- artefact delivered but incorrect.} Artefacts complete, requirement covered --- and a number, a matrix or a balance that the verifier recomputes differently. It reported a value it never checked.
\emph{Prescription:} add an ending verification step: re-read the artefact from disk, recompute the reported quantity by a second route, compare against an explicit threshold, and loop until it holds. ``I wrote the file'' and ``the run printed no error'' are not evidence.
\end{enumerate}

\textbf{Rules}
\begin{itemize}[nosep, leftmargin=1.4em]
\item Edit what the evidence implicates; there is no required document shape, and restructuring the whole file is never required by this guide.
\item Keep documented capability the task might need; cut only what the agent cannot act on.
\item Only mention functions, flags and file paths present in the supplied task, skill files, trajectory, verifier output, or executor interface. Do not invent names.
\item Do not paste expected outputs, expected numbers, or a solution. A skill that works only because it contains the answer is rejected.
\item If the evidence does not support a grounded improvement, return NO\_CHANGE.
\end{itemize}

Follow the output contract in the user message: the complete rewritten file for every file you change, or a single NO\_CHANGE block.
\end{promptbox}

\begin{promptbox}{No-meta baseline (system prompt of the \emph{w/o meta-skill} arm)}
\textbf{Skill rewrite (baseline, no meta-skill)}

You are editing one \textbf{skill} --- a document that is injected into a coding agent's context before it starts a task. The agent that will read your output is a small, weak model. It has already tried this task once with the current version of the skill. You are given what happened. Produce a better version.

The skill is the only thing you may change. You cannot change the task, the agent, or the tests.

Follow the output contract in the user message: return the complete rewritten file for every file you change, or a single NO\_CHANGE block. Two hard rules:
\begin{itemize}[nosep, leftmargin=1.4em]
\item The frontmatter must survive with the same \texttt{name}.
\item Do not paste expected outputs, expected numbers, or a solution. A skill that works only because it contains the answer is rejected.
\end{itemize}
\end{promptbox}

\begin{promptbox}{One-sentence instruction (skill adaptation comparison, \cref{fig:no-meta})}
\textbf{Skill Adaptation (plain)}

Rewrite this skill so that it works for all models.

Do not paste expected outputs, expected numbers, or a solution. If no grounded improvement is possible, return NO\_CHANGE.

Follow the output contract in the user message: the complete rewritten file for every file you change, or a single NO\_CHANGE block.
\end{promptbox}

\section{Full Self-Evolution Results}
\label{app:rsi}
\begin{figure}[h]
\centering
\includegraphics[width=\textwidth]{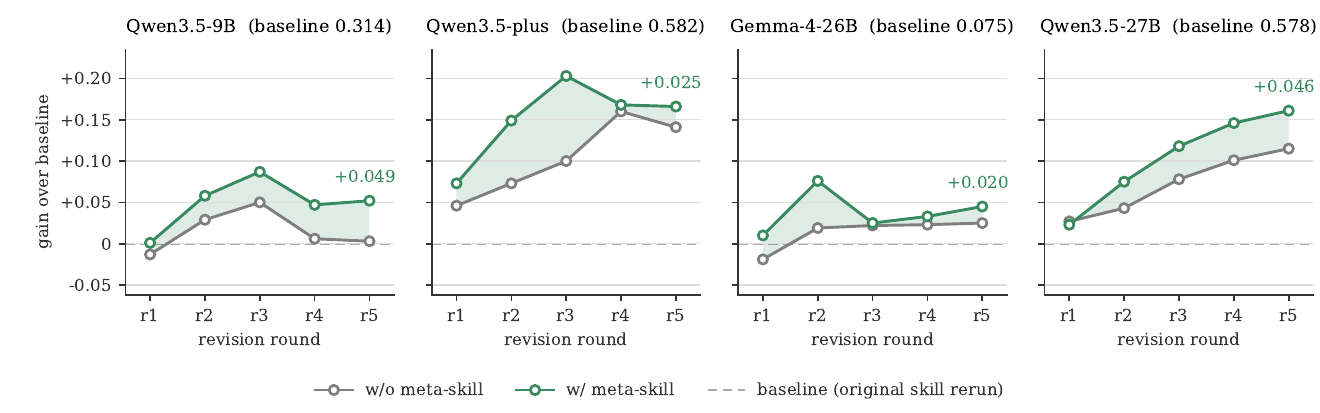}
\caption{Self-evolution with and without the meta-skill on the four executors not shown in \cref{fig:rsi}. Curves show the gain of each round over the baseline that reruns the original skill; annotations mark the gap between the two arms at round~5.}
\label{fig:rsi-appendix}
\end{figure}

\cref{tab:rsi-full} lists the mean score of every executor in every round of the self-evolution experiment (\cref{subsec:rsi}), with and without the meta-skill, and \cref{fig:rsi-appendix} plots the four executors not shown in \cref{fig:rsi}.
The protocol is identical for all seven executors: each rewrites its own skill from the previous round's trajectory and verifier output over five rounds, every round is run once on the $103$ pairs, and the baseline reruns the original skill in the same time window.
On all seven executors the arm with the meta-skill ends above the arm without it at round~5, by $+0.020$ to $+0.051$, although neither arm improves monotonically and the two arms cross on several executors in the early rounds.

\begin{table}[h]
\centering
\caption{Mean score in each round of self-evolution, without (w/o) and with (w/) the meta-skill, overall and on each benchmark. \emph{Base} reruns the original skill in the same time window; $\Delta_5$ is the gap between the two arms at round~5, positive when the meta-skill arm is higher.}
\label{tab:rsi-full}
\scriptsize
\setlength{\tabcolsep}{4.2pt}
\renewcommand{\arraystretch}{1.12}
\begin{tabular}{l l c ccccc >{\columncolor{metatint}}c >{\columncolor{metatint}}c >{\columncolor{metatint}}c >{\columncolor{metatint}}c >{\columncolor{metatint}}c c}
\toprule
\multirow{2}{*}{Executor} & \multirow{2}{*}{Benchmark} & \multirow{2}{*}{Base} & \multicolumn{5}{c}{w/o meta-skill} & \multicolumn{5}{c}{\cellcolor{metatint}w/ meta-skill} & \multirow{2}{*}{$\Delta_5$} \\
\cmidrule(lr){4-8}\cmidrule(lr){9-13}
 & & & $r_1$ & $r_2$ & $r_3$ & $r_4$ & $r_5$ & $r_1$ & $r_2$ & $r_3$ & $r_4$ & $r_5$ & \\
\midrule
\rowcolor{grouptint}\multicolumn{14}{l}{\emph{Weak executors used to build the meta-skill (\cref{subsec:rubric})}} \\
\multirow{4}{*}{\micon{qwen}Qwen3.6-flash} & \textbf{All} & \textbf{0.523} & \textbf{0.545} & \textbf{0.549} & \textbf{0.611} & \textbf{0.599} & \textbf{0.614} & \textbf{0.567} & \textbf{0.613} & \textbf{0.616} & \textbf{0.621} & \textbf{0.644} & \gain{+0.030} \\
 & SkillsBench & 0.350 & 0.309 & 0.239 & 0.412 & 0.293 & 0.356 & 0.337 & 0.397 & 0.343 & 0.313 & 0.417 & \gain{+0.061} \\
 & DocOps & 0.375 & 0.438 & 0.438 & 0.469 & 0.625 & 0.531 & 0.375 & 0.469 & 0.531 & 0.531 & 0.500 & \loss{-0.031} \\
 & Tessl & 0.786 & 0.903 & 0.895 & 0.959 & 0.946 & 0.893 & 0.912 & 0.908 & 0.909 & 0.948 & 0.947 & \gain{+0.054} \\
\cmidrule(lr){2-14}
\multirow{4}{*}{\micon{gemma}Gemma-4-E4B} & \textbf{All} & \textbf{0.104} & \textbf{0.097} & \textbf{0.060} & \textbf{0.089} & \textbf{0.071} & \textbf{0.096} & \textbf{0.109} & \textbf{0.092} & \textbf{0.116} & \textbf{0.161} & \textbf{0.147} & \gain{+0.051} \\
 & SkillsBench & 0.074 & 0.020 & 0.046 & 0.038 & 0.042 & 0.031 & 0.033 & 0.067 & 0.008 & 0.078 & 0.043 & \gain{+0.012} \\
 & DocOps & 0.000 & 0.062 & 0.000 & 0.000 & 0.000 & 0.000 & 0.000 & 0.000 & 0.031 & 0.000 & 0.031 & \gain{+0.031} \\
 & Tessl & 0.215 & 0.259 & 0.207 & 0.205 & 0.153 & 0.227 & 0.262 & 0.189 & 0.275 & 0.362 & 0.326 & \gain{+0.099} \\
\cmidrule(lr){2-14}
\multirow{4}{*}{\micon{qwen}Qwen3.5-9B} & \textbf{All} & \textbf{0.314} & \textbf{0.301} & \textbf{0.343} & \textbf{0.364} & \textbf{0.320} & \textbf{0.317} & \textbf{0.315} & \textbf{0.372} & \textbf{0.401} & \textbf{0.361} & \textbf{0.366} & \gain{+0.049} \\
 & SkillsBench & 0.112 & 0.250 & 0.146 & 0.203 & 0.112 & 0.138 & 0.201 & 0.095 & 0.149 & 0.157 & 0.158 & \gain{+0.020} \\
 & DocOps & 0.156 & 0.156 & 0.250 & 0.250 & 0.281 & 0.125 & 0.125 & 0.250 & 0.281 & 0.219 & 0.281 & \gain{+0.156} \\
 & Tessl & 0.610 & 0.522 & 0.627 & 0.591 & 0.684 & 0.622 & 0.563 & 0.546 & 0.706 & 0.646 & 0.605 & \loss{-0.017} \\
\midrule
\rowcolor{grouptint}\multicolumn{14}{l}{\emph{Additional executors}} \\
\multirow{4}{*}{\micon{qwen}Qwen3.5-27B} & \textbf{All} & \textbf{0.578} & \textbf{0.605} & \textbf{0.621} & \textbf{0.656} & \textbf{0.679} & \textbf{0.693} & \textbf{0.601} & \textbf{0.653} & \textbf{0.696} & \textbf{0.724} & \textbf{0.739} & \gain{+0.046} \\
 & SkillsBench & 0.416 & 0.422 & 0.395 & 0.617 & 0.512 & 0.460 & 0.397 & 0.404 & 0.421 & 0.557 & 0.503 & \gain{+0.043} \\
 & DocOps & 0.438 & 0.531 & 0.469 & 0.562 & 0.531 & 0.625 & 0.438 & 0.562 & 0.469 & 0.594 & 0.688 & \gain{+0.063} \\
 & Tessl & 0.827 & 0.911 & 0.932 & 0.939 & 0.939 & 0.939 & 0.869 & 0.933 & 0.939 & 0.968 & 0.975 & \gain{+0.036} \\
\cmidrule(lr){2-14}
\multirow{4}{*}{\micon{gemma}Gemma-4-26B} & \textbf{All} & \textbf{0.075} & \textbf{0.056} & \textbf{0.094} & \textbf{0.097} & \textbf{0.098} & \textbf{0.100} & \textbf{0.085} & \textbf{0.151} & \textbf{0.100} & \textbf{0.108} & \textbf{0.120} & \gain{+0.020} \\
 & SkillsBench & 0.062 & 0.044 & 0.031 & 0.038 & 0.034 & 0.045 & 0.018 & 0.021 & 0.039 & 0.039 & 0.108 & \gain{+0.063} \\
 & DocOps & 0.000 & 0.000 & 0.000 & 0.031 & 0.031 & 0.031 & 0.031 & 0.000 & 0.000 & 0.000 & 0.000 & \loss{-0.031} \\
 & Tessl & 0.147 & 0.229 & 0.205 & 0.199 & 0.204 & 0.203 & 0.183 & 0.185 & 0.233 & 0.252 & 0.228 & \gain{+0.025} \\
\cmidrule(lr){2-14}
\multirow{4}{*}{\micon{qwen}Qwen3.5-plus} & \textbf{All} & \textbf{0.582} & \textbf{0.628} & \textbf{0.655} & \textbf{0.682} & \textbf{0.742} & \textbf{0.723} & \textbf{0.655} & \textbf{0.731} & \textbf{0.785} & \textbf{0.750} & \textbf{0.748} & \gain{+0.025} \\
 & SkillsBench & 0.502 & 0.507 & 0.357 & 0.487 & 0.537 & 0.446 & 0.559 & 0.521 & 0.407 & 0.575 & 0.667 & \gain{+0.221} \\
 & DocOps & 0.375 & 0.469 & 0.594 & 0.656 & 0.781 & 0.719 & 0.406 & 0.688 & 0.562 & 0.656 & 0.531 & \loss{-0.188} \\
 & Tessl & 0.817 & 0.899 & 0.948 & 0.935 & 0.952 & 0.955 & 0.850 & 0.938 & 0.956 & 0.969 & 0.992 & \gain{+0.037} \\
\cmidrule(lr){2-14}
\multirow{4}{*}{\micon{kimi}Kimi-K2.5} & \textbf{All} & \textbf{0.612} & \textbf{0.683} & \textbf{0.663} & \textbf{0.682} & \textbf{0.702} & \textbf{0.765} & \textbf{0.690} & \textbf{0.694} & \textbf{0.764} & \textbf{0.760} & \textbf{0.800} & \gain{+0.035} \\
 & SkillsBench & 0.474 & 0.568 & 0.519 & 0.498 & 0.621 & 0.604 & 0.460 & 0.499 & 0.671 & 0.476 & 0.644 & \gain{+0.040} \\
 & DocOps & 0.438 & 0.469 & 0.469 & 0.562 & 0.688 & 0.656 & 0.469 & 0.531 & 0.625 & 0.531 & 0.750 & \gain{+0.094} \\
 & Tessl & 0.856 & 0.912 & 0.925 & 0.913 & 0.970 & 0.970 & 0.905 & 0.967 & 0.945 & 0.956 & 0.953 & \loss{-0.017} \\
\bottomrule
\end{tabular}
\end{table}

\section{Additional Case Studies}
\label{app:cases}
We complement \cref{fig:case} with four further pairs on which the two arms differ clearly: two from self-evolution, a DocOps spreadsheet task by Qwen3.5-9B (\cref{fig:case-excel}) and a Tessl developer task by Gemma-4-E4B (\cref{fig:case-stripe}), and two from one-shot rewriting by Kimi-K3, a SkillsBench planning task run by Gemma-4-E4B (\cref{fig:case-pddl}) and a SkillsBench bioinformatics task run by Qwen3.6-flash (\cref{fig:case-protein}).
Each figure follows the layout of \cref{fig:case}: the left box shows the skill produced without the meta-skill, the right box the skill produced with it, passages aligned by the failure mode they address, and the reward line gives the task reward per round.
All quotes are taken from the skills and trajectories on disk; as every cell is a single rollout, the cases illustrate the mechanisms behind \cref{tab:rubric} rather than prove them; the aggregate evidence is the table itself.

\paragraph{Spreadsheet totals (DocOps, self-evolution).}
\cref{fig:case-excel}. The task asks for the Maintenance and Operations totals, computed by formulas from a cost-code mapping, in the summary area of a \texttt{Cost Tracker} sheet, saved to a submission path.
The original \texttt{xlsx} skill is an $11.5$k-character library reference that never mentions the task; Qwen3.5-9B fails in every round without the meta-skill, and in rounds 1--3 with it, always by writing correct or nearly correct formulas into the wrong cells, e.g.\ overwriting the label in H3 or placing the total in J3.
Without the meta-skill the rewriter chases symptoms: it adds Python syntax tips, misdiagnoses the recalculation script as the culprit and forbids running it, and in round~3 hard-codes the wrong geometry (``place formulas below the labels''), on which later rounds build; in round~4 the executor stops without writing any file.
With the meta-skill, round~3 fixes the computation and round~4 adds the two passages that make the difference: formula cells are distinguished from header cells, and the workbook is read back and compared against totals recomputed from the source rows.
The passing skill is task-specific (it names I3 and I4), consistent with our reading of self-evolution as per-task adaptation.

\begin{figure}[t]
\centering
\begin{tcbraster}[raster columns=2, raster equal height, raster column skip=6pt, raster valign=top]
\begin{casebox}{\casetitle{w/o meta-skill}{skill after round 5}}
\begin{plainbox}{\badmark}\caseh{Task \& Delivery} Never stated: the target cells, the output path and the mapping sheet are left to the executor.\end{plainbox}
\begin{plainbox}{\badmark}\caseh{Recalculation} ``Do NOT run \texttt{recalc.py}'': a false diagnosis that removes the only check on the totals.\end{plainbox}
\begin{plainbox}{\badmark}\caseh{Placement} ``Place formulas below the labels'': the opposite of the sheet's layout, kept in every later round.\end{plainbox}
\begin{plainbox}{\badmark}\caseh{Checklists} Syntax tips and a five-step workflow checklist; nothing names a cell or compares a value.\end{plainbox}
\tcblower
\casereward{\badmark}{\badmark\,\badmark\,\badmark\,\badmark\,\shown{\badmark}}{Wrong in every round: the skill grows by $66\%$ yet the totals never land in I3 and I4; round~4 ends with no file written at all.}
\end{casebox}
\begin{metacasebox}{\casetitle{w/ meta-skill}{skill after round 4}}
\begin{metabox}{Context management}{\okmark}\caseh{Delivery Path} ``Save to the correct output path as specified in the task'' as an explicit step.\end{metabox}
\begin{metabox}{Instruction following}{\okmark}\caseh{Formula vs.\ Header Cell} ``Formulas go in the adjacent column, not the same column as the header''; I3 and I4 named as targets.\end{metabox}
\begin{metabox}{Verification}{\okmark}\caseh{Computation} \texttt{=SUMIF(C4:C14,"M-*",D4:D14)} over the cost-code column rather than the blank bucket column.\end{metabox}
\begin{metabox}{Verification}{\okmark}\caseh{Read-back} ``Load the file with \texttt{data\_only=False}, check that the formula cells start with \texttt{=}'' before saving.\end{metabox}
\tcblower
\casereward{\badmark}{\badmark\,\badmark\,\badmark\,\shown{\okmark}\,\okmark}{Passes from round 4: the executor prints the summary block with column letters, writes I3 and I4, and compares the read-back totals with sums recomputed from the source rows.}
\end{metacasebox}
\end{tcbraster}
\caption{Self-evolution by Qwen3.5-9B on a DocOps spreadsheet task. Marks give the task reward from round~0 to round~5; the framed mark is the round shown.}
\label{fig:case-excel}
\end{figure}

\paragraph{Project setup files (Tessl, self-evolution).}
\cref{fig:case-stripe}. The user starts a React/Node project on a Mac and asks for two files, a \texttt{setup.sh} that installs the Stripe Projects CLI and a \texttt{STRIPE\_SETUP.md} for team members.
The original skill is five lines written for an agent that will run the CLI interactively; it never says that the deliverable is two written files.
Gemma-4-E4B's characteristic failure is a tool execution failure: it mistypes the skill name, receives \texttt{Unknown skill}, relaunches the skill under a valid name, which only re-injects the same text, and ends its turn announcing that the files will follow, without ever calling \texttt{Write}.
Without the meta-skill the rewriter transcribes the judge's rubric into a checklist of properties the files must have, then returns \texttt{NO\_CHANGE} or a byte-identical rewrite after each failure; the checklist says nothing about what is callable or where the work ends, and the executor fails in three of five rounds.
With the meta-skill the rewriter applies guidance for context management in round~1 and tool execution in round~2, adding a task scope, a delivery path and the install commands one per line; it then returns \texttt{NO\_CHANGE} for rounds 3--5, and the executor passes every round, including two rounds in which it hits the same \texttt{Unknown skill} error but now falls through to \texttt{Write} instead of stopping.

\begin{figure}[t]
\centering
\begin{tcbraster}[raster columns=2, raster equal height, raster column skip=6pt, raster valign=top]
\begin{casebox}{\casetitle{w/o meta-skill}{skill after round 5}}
\begin{plainbox}{\badmark}\caseh{Requirements Checklist} ``Must contain at least three non-comment shell commands''; ``at least one heading'': the judge's rubric restated as properties.\end{plainbox}
\begin{plainbox}{\badmark}\caseh{Callable Surface} Nothing says that launching a skill produces no file and that the executor must call \texttt{Write} itself.\end{plainbox}
\begin{plainbox}{\badmark}\caseh{Commands} The verbatim \texttt{brew install} line is dropped for ``e.g., using Homebrew''.\end{plainbox}
\begin{plainbox}{\badmark}\caseh{Later Rounds} \texttt{NO\_CHANGE} after round~1 and a byte-identical rewrite after round~4, despite two failing trajectories in between.\end{plainbox}
\tcblower
\casereward{\badmark}{\okmark\,\badmark\,\okmark\,\badmark\,\shown{\badmark}}{Passes only when the first skill launch succeeds; whenever the name is mistyped, the executor relaunches the skill and stops without writing.}
\end{casebox}
\begin{metacasebox}{\casetitle{w/ meta-skill}{skill after round 2}}
\begin{metabox}{Context management}{\okmark}\caseh{Task Scope} ``The agent must write two files: \texttt{setup.sh} \ldots\ and \texttt{STRIPE\_SETUP.md}''.\end{metabox}
\begin{metabox}{Context management}{\okmark}\caseh{Delivery Path} ``The agent must write these two files to the workspace.''\end{metabox}
\begin{metabox}{Tool execution}{\okmark}\caseh{Setup Instructions} Install and init commands one per line, verbatim: \texttt{brew install stripe/stripe-cli/stripe}, \texttt{stripe projects init}.\end{metabox}
\begin{metabox}{Instruction following}{\okmark}\caseh{Completion} ``The \texttt{setup.sh} script must wrap these steps. The \texttt{STRIPE\_SETUP.md} must document this entire process.''\end{metabox}
\tcblower
\casereward{\badmark}{\okmark\,\shown{\okmark}\,\okmark\,\okmark\,\okmark}{Passes every round; rounds 3--5 return \texttt{NO\_CHANGE}. When the skill launch fails, the scope and delivery lines carry the executor to \texttt{Write}.}
\end{metacasebox}
\end{tcbraster}
\caption{Self-evolution by Gemma-4-E4B on a Tessl project-setup task. The skill with the meta-skill is unchanged after round~2, the rewriter returning \texttt{NO\_CHANGE} in rounds 3--5.}
\label{fig:case-stripe}
\end{figure}

\paragraph{PDDL planning (SkillsBench, one-shot rewriting).}
\cref{fig:case-pddl}. For each entry of \texttt{problem.json}, the executor must generate a valid sequential plan for a PDDL domain and problem and write it, one action per line, to the entry's \texttt{plan\_output} path.
The original $3.1$k-character skill is an API reference for four pseudo-functions (\texttt{load\_problem}, \texttt{generate\_plan}, \texttt{save\_plan}, \texttt{validate}) that exist neither as tools nor as an installed module; Gemma-4-E4B treats them as callables, derives a plan by hand and ends with ``the plan is ready for manual saving'', writing nothing.
Both Kimi-K3 rewrites replace the document and both state the deliverable and that the functions are not tools, so the difference lies in the operational detail.
The one-sentence rewrite opens with a six-line block of chained optional \texttt{pip install}s; the executor fuses it into one command with a stray quote, the persistent terminal hangs, and the fifteen remaining calls, including a heredoc that was meant to write the script, all return ``process still running''.
The meta-skill rewrite gates the preflight on a single probe (``do not proceed until it prints \texttt{up ok}''), names \texttt{file\_editor} as the tool that writes the script, gives every entry an action, target and completion condition, and ends with a disk check that must print one \texttt{OK} per entry; the executor meets an import error, follows the fallback, removes the offending import and loops until the disk check prints \texttt{OK} for both entries.

\begin{figure}[t]
\centering
\begin{tcbraster}[raster columns=2, raster equal height, raster column skip=6pt, raster valign=top]
\begin{casebox}{\casetitle{one-sentence instruction}{Kimi-K3 rewrite}}
\begin{plainbox}{\okmark}\caseh{Deliverable} Stated up front: load, plan, ``write the plan to the exact path given by \texttt{plan\_output}''.\end{plainbox}
\begin{plainbox}{\badmark}\caseh{Setup} Six chained lines, \texttt{pip install \ldots\ || pip install \ldots\ || true}, with no pass/fail gate; the executor fuses them and hangs the terminal.\end{plainbox}
\begin{plainbox}{\badmark}\caseh{Script} ``Write one script'' without naming \texttt{file\_editor}; the executor writes it through the hung terminal.\end{plainbox}
\begin{plainbox}{\badmark}\caseh{Verification} ``\texttt{ls -la}'' and ``\texttt{head -3}'' placeholders with no completion condition; an empty plan file is kept as a ``last resort''.\end{plainbox}
\tcblower
{\footnotesize\textbf{Reward:}~original~\badmark~~$\vert$~~rewritten~\badmark\quad The right content, wrong operational form: one ungated command locks the only channel the skill told the executor to use, and nothing pulls it back to the artefact.}
\end{casebox}
\begin{metacasebox}{\casetitle{w/ meta-skill}{Kimi-K3 rewrite}}
\begin{metabox}{Context management}{\okmark}\caseh{Task First} ``For every entry \ldots'' of \texttt{problem.json}; ``everything after this section is reference material, not a new request.''\end{metabox}
\begin{metabox}{Tool execution}{\okmark}\caseh{Callable Surface} ``\texttt{load\_problem} \ldots\ are Python library calls inside a script; they are not tools. Your only callable tools are \texttt{terminal}, \texttt{file\_editor}, \ldots''\end{metabox}
\begin{metabox}{Tool execution}{\okmark}\caseh{Gated Preflight} One import probe; ``do not proceed until it prints \texttt{up ok}''; set the terminal \texttt{timeout} field.\end{metabox}
\begin{metabox}{Instruction following}{\okmark}\caseh{Requirement Checklist} ``Action: write the plan --- Target: exactly the \texttt{plan\_output} path --- Completion: a file exists at that path''.\end{metabox}
\begin{metabox}{Verification}{\okmark}\caseh{Verify from Disk} ``A printed \texttt{WROTE} line is necessary but not sufficient''; loop until exactly one \texttt{OK} per entry.\end{metabox}
\tcblower
{\footnotesize\textbf{Reward:}~original~\badmark~~$\vert$~~rewritten~\okmark\quad The executor hits an import error, follows the fallback, edits the script twice and stops only when the disk check prints \texttt{OK} for both entries.}
\end{metacasebox}
\end{tcbraster}
\caption{One-shot rewriting by Kimi-K3 of a SkillsBench PDDL planning skill, run by Gemma-4-E4B. Both rewrites state the task; only the meta-skill rewrite gives the executor a gated preflight, a named writing tool and a completion condition for every entry.}
\label{fig:case-pddl}
\end{figure}

\paragraph{Protein expression workbook (SkillsBench, one-shot rewriting).}
\cref{fig:case-protein}. The executor must fill the \texttt{Task} sheet of a workbook with live formulas: a two-way lookup of ten proteins by ten samples, per-group means and standard deviations, and log fold changes; the verifier reads cached values only, so formulas that are never recalculated score zero.
This is a case where the original $10.6$k-character \texttt{xlsx} skill already works for Qwen3.6-flash, and the question is whether a rewrite preserves that.
Both rewrites shorten the skill, by $23\%$ and $30\%$, and both keep the recalculation step, but the original gives the command as \texttt{python recalc.py} and neither the original nor the one-sentence rewrite says what to do if \texttt{python} is absent.
With the one-sentence rewrite, the executor writes all $160$ formulas, runs the literal command, receives \texttt{python: command not found}, and ends its turn; the verifier finds no cached values and fails three of six tests.
The meta-skill rewrite adds one preflight line, ``use \texttt{python3} if \texttt{python} is missing'', and a mandatory final step that re-opens the saved file and cross-checks it against the raw data with pandas.
The same error occurs, the executor retries with \texttt{python3}, and the cross-check then catches a lookup range that was offset by two columns, which the recalculation script had passed with zero errors; two fix--recalculate--verify iterations later all six tests pass, including the three that the one-sentence rewrite failed.

\begin{figure}[t]
\centering
\begin{tcbraster}[raster columns=2, raster equal height, raster column skip=6pt, raster valign=top]
\begin{casebox}{\casetitle{one-sentence instruction}{Kimi-K3 rewrite}}
\begin{plainbox}{\badmark}\caseh{Generic Manual} A tidy workflow for \texttt{file.xlsx} and \texttt{SheetName}; the task, its ranges and its workbook are never named.\end{plainbox}
\begin{plainbox}{\badmark}\caseh{Recalculate} ``Locate \texttt{recalc.py} with \texttt{find}, then \texttt{python recalc.py file.xlsx}'': the literal interpreter, no fallback.\end{plainbox}
\begin{plainbox}{\okmark}\caseh{Verify} ``Reload with \texttt{data\_only=True}: every target cell must hold a computed value'', never reached.\end{plainbox}
\tcblower
{\footnotesize\textbf{Reward:}~original~\okmark~~$\vert$~~rewritten~\badmark\quad A working skill is broken: the rewrite inherits the latent \texttt{python} hazard and makes the exact failing command more prominent; the executor stops at the error with $160$ unrecalculated formulas.}
\end{casebox}
\begin{metacasebox}{\casetitle{w/ meta-skill}{Kimi-K3 rewrite}}
\begin{metabox}{Context management}{\okmark}\caseh{Current Task} ``You are completing the differential-expression worksheet in \texttt{protein\_expression.xlsx}''; ``edited in place, no new files, no new sheets.''\end{metabox}
\begin{metabox}{Tool execution}{\okmark}\caseh{Preflight} \texttt{python -c "import openpyxl, pandas"} --- ``use \texttt{python3} if \texttt{python} is missing''; locate \texttt{recalc.py} with \texttt{find}.\end{metabox}
\begin{metabox}{Instruction following}{\okmark}\caseh{One Requirement at a Time} ``\texttt{C11:L20} --- two-way lookup \ldots\ Completion: all 100 cells hold formulas''; likewise for \texttt{B24:K27} and \texttt{C32:D41}.\end{metabox}
\begin{metabox}{Verification}{\okmark}\caseh{Verify from Disk} ``\,`Saved without error' is not evidence the numbers are right'': reopen with \texttt{data\_only=True}, cross-check with pandas, repeat until every check passes.\end{metabox}
\tcblower
{\footnotesize\textbf{Reward:}~original~\okmark~~$\vert$~~rewritten~\okmark\quad The same \texttt{python} error occurs; the executor retries with \texttt{python3}, and the cross-check catches an offset lookup range that recalculation had passed, closing after two fix--verify loops.}
\end{metacasebox}
\end{tcbraster}
\caption{One-shot rewriting by Kimi-K3 of a SkillsBench spreadsheet skill, run by Qwen3.6-flash. The one-sentence rewrite breaks a skill that worked; the meta-skill rewrite keeps it working and adds a check that catches a wrong value the recalculation script had passed.}
\label{fig:case-protein}
\end{figure}

\end{document}